\documentclass[sn-mathphys,Numbered]{sn-jnl}% Math and Physical Sciences Reference Style
\usepackage{graphicx}%
\usepackage{multirow}%
\usepackage{amsmath,amssymb,amsfonts}%
\usepackage{amsthm}%
\usepackage{mathrsfs}%
\usepackage[title]{appendix}%
\usepackage{xcolor}%
\usepackage{textcomp}%
\usepackage{manyfoot}%
\usepackage{booktabs}%
\usepackage{algorithm}%
\usepackage{algorithmicx}%
\usepackage{algpseudocode}%
\usepackage{listings}%
\usepackage{hyperref}
\usepackage{xcolor}

\newcommand{\MV}[1]{\textcolor{black}{#1}}

\usepackage{wrapfig}
\usepackage{xcolor}
\usepackage{lipsum}% dummy text only
\usepackage{caption}
\usepackage{capt-of}
\usepackage{subcaption}
\usepackage{rotating}
\usepackage{comment}

\usepackage{tabularx}
\usepackage{adjustbox}

\theoremstyle{thmstyleone}%
\theoremstyle{thmstyletwo}%

\theoremstyle{thmstylethree}%

\begin{document}

\title[Article Title]{Artificial Intelligence for Geometry-Based Feature Extraction, Analysis and Synthesis in \MV{Artistic Images}: A Survey}


\author[1]{\fnm{Mridula} \sur{Vijendran}}\email{mridula.vijendran@durham.ac.uk}

\author[1]{\fnm{Jingjing} \sur{Deng}}\email{jingjing.deng@durham.ac.uk}
\author[1]{\fnm{Shuang} \sur{Chen}}\email{shuang.chen@durham.ac.uk}

\author[2]{\fnm{Edmond S. L.} \sur{Ho}}\email{shu-lim.ho@glasgow.ac.uk}

\author*[1]{\fnm{Hubert P. H.} \sur{Shum}}\email{hubert.shum@durham.ac.uk}

\affil[1]{\orgdiv{Department of Computer Science}, \orgname{Durham University}, \orgaddress{\city{Durham}, \\ \postcode{DH1 3LE}, \country{United Kingdom}}}
\affil[2]{\orgdiv{School of Computer Science}, \orgname{University of Glasgow}, \orgaddress{\city{Glasgow}, \\ \postcode{G12 8RZ}, \country{United Kingdom}}}


\abstract{Artificial Intelligence significantly enhances the visual art industry by analyzing, identifying and generating digitized \MV{artistic images}. This review highlights the substantial benefits of integrating geometric data into AI models, addressing challenges such as high inter-class variations, domain gaps, and the separation of style from content by incorporating geometric information. Models not only improve \MV{AI-generated graphics} synthesis quality, but also effectively distinguish between style and content, utilizing inherent model biases and shared data traits. We explore methods like geometric data extraction from \MV{artistic images}, the impact on human perception, and its use in discriminative tasks. The review also discusses the potential for improving data quality through innovative annotation techniques and the use of geometric data to enhance model adaptability and output refinement. Overall, incorporating geometric guidance boosts model performance in classification and synthesis tasks, providing crucial insights for future AI applications in the \MV{visual arts domain}.}




\maketitle

\begingroup
\let\clearpage\relax

\section{Introduction}\label{sec1}
\begin{figure}

    \centering
    \includegraphics[width=0.7\linewidth,]{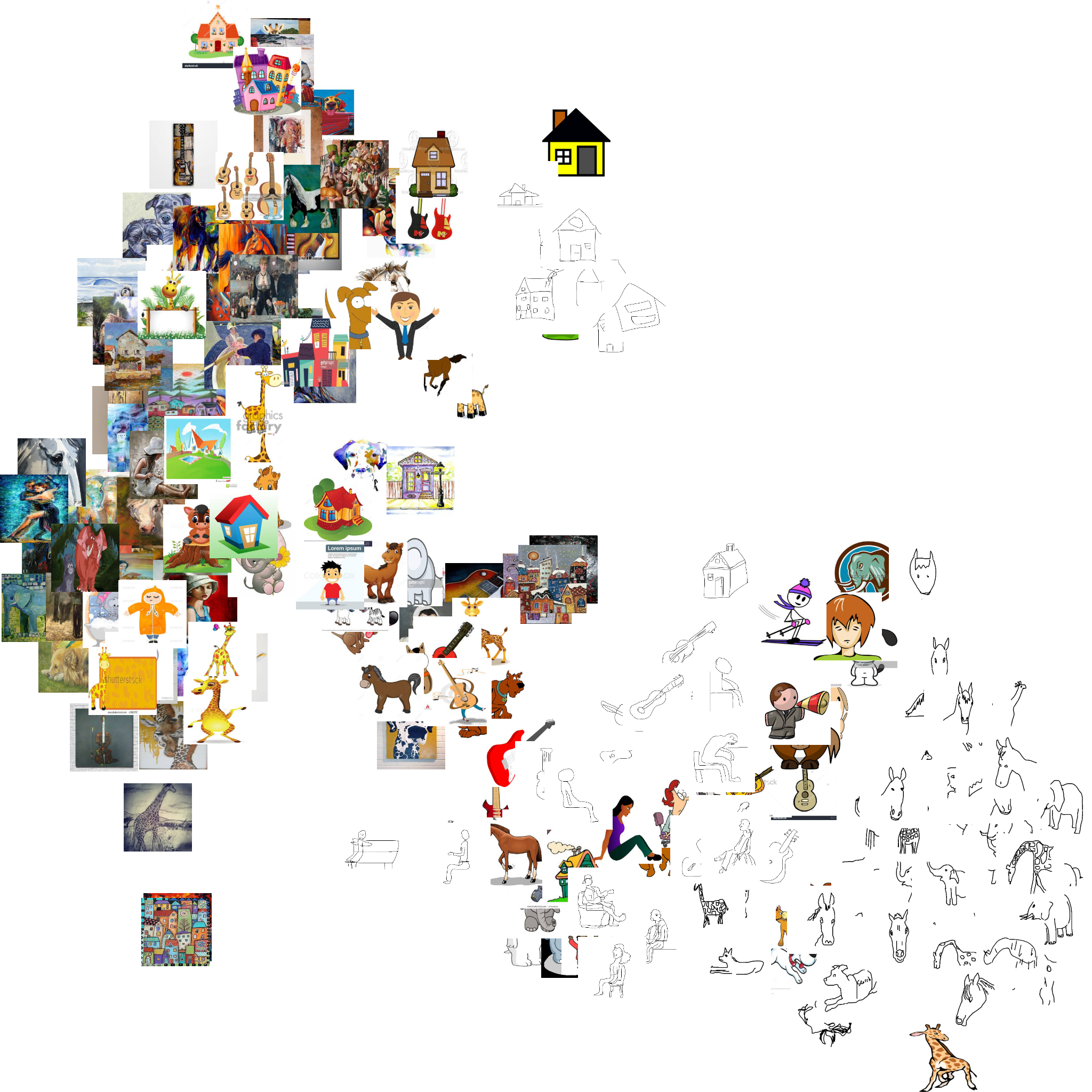}
    \includegraphics[width=0.5\linewidth]{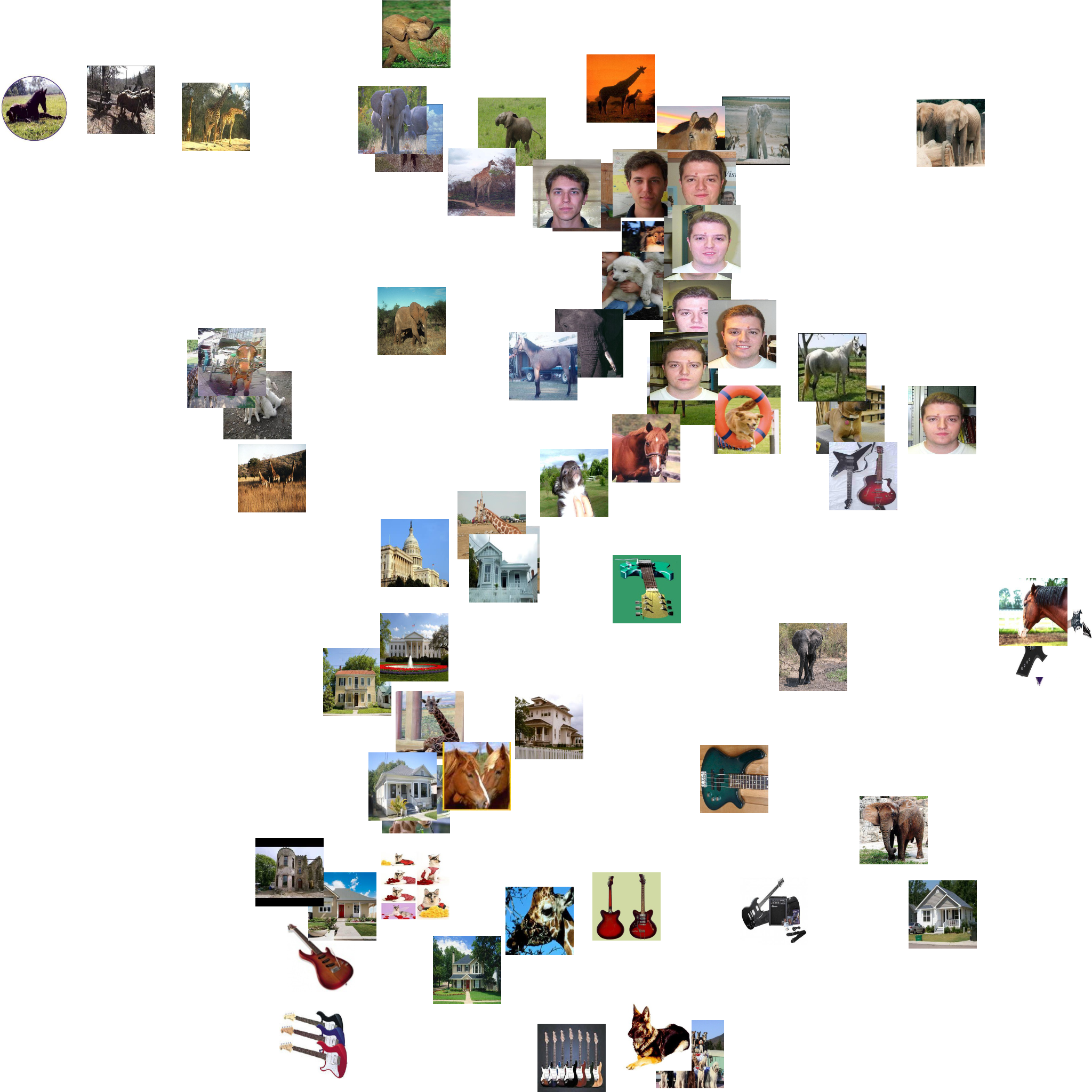}
    \caption{T-SNE visualization of the art domain as compared to the real-world images using the PACS dataset. The art modalities with paintings, cartoons and sketches showcase the clustering of art modalities close to photos, exaggerated geometries and no color or texture respectively. The dimensionality reduction uses a pre-trained VGG-19 model as a feature extractor and removes the fully connected head. }\label{fig:tsne_vis}
    
\end{figure}
Artificial intelligence (AI) techniques find use in the art industry for tasks such as 3D scan analysis, art recommendation systems, identification of art design principles, deconstructivism art generation with fragment models \cite{bellaiche2023humans,hirsch2021practice}. These techniques mainly involve three key processes: extraction, analysis, and synthesis. Extraction methods classify paintings based on style, identify and authenticate artwork, and provide exhibit and tour information to establishments like museums \cite{rani2023exploring} and historic cathedrals \cite{Sklodowski2014}  to enhance their visitors’ experience. Analysis techniques interpret and facilitate searching and comparing art elements such as geometric patterns of compositional elements across multiple scales in art collections \cite{Pintus2016}. Synthesis methods are used for scanning and enhancing the details of artifacts in the Cultural Heritage field \cite{borg2020application, Remondino2011} and deal with the preservation, documentation and collection of historical and cultural objects.
They classify paintings based on style, identify and authenticate artwork, and provide virtual access to historic cathedrals to enhance their consumers' experience. It is used for scanning and enhancing the details of artifacts in the Cultural Heritage field \cite{borg2020application, pintus2016survey, Remondino2011} which deals with the preservation, documentation and collection of historical and cultural objects.

This paper explores a broad spectrum of AI techniques applied to artworks, encompassing both deep learning and non-deep learning methods. While deep learning models, such as CNNs and GANs, have shown significant success in tasks like object detection and image synthesis, we also discuss traditional methods, including Deformable Part Models (DPM), Histogram of Oriented Gradients (HoG), and Thin Plate Spline (TPS) interpolation. These non-deep learning techniques are crucial for understanding specific geometric features and enhancing overall model performance when integrated with deep learning approaches.

Learning from Art datasets using artificial intelligence models is challenging due to the generally smaller dataset size and larger inter-class variations \cite{redmon2016you, Mathieu2014}, as well as incomplete or inaccurate data annotations \cite{Milani2022, Elliot2013}. This review paper broadly covers 3D art such as sculptures, archaeological sites and surface art such as walls, cloth or tattoos. Additionally, it also considers 2D forms such as paintings, sketches, digital art, cartoons, logos, anime and manga. Synthetic art through the stylization of real-world data without content separation commonly suffers from the bleeding of colors from the foreground to the background and the blurring of boundaries. Despite depicting the same content, the stylistic differences between various art media highlight the importance of separating style and content in model design for different art-related tasks. To illustrate these problems, we use t-distributed Stochastic Neighbor Embedding (T-SNE), which is a commonly used nonlinear dimensionality reduction algorithm to visualize embeddings and understand similar images from the dataset according to the model, VGG-19, embedding the data. From the T-SNE visualization in Figure \ref{fig:tsne_vis}, we see that art domains of one type cluster closely, with some overlap for those sharing similar shape representations with cartoons and sketches. The large inter-class variations in the art modalities lead to their distributions looking roughly like anisotropic Gaussians, with their spread creating overlaps with images of similar subjects between the other classes. The stylization of real-world data without content separation leads to the bleeding of colors from the foreground to the background and the blurring of boundaries. Our paper discusses paintings, sketches, digital art, cartoons, logos, anime and manga as visual art media. It covers a variety of styles and genres such as abstract art, realism, impressionism, expressionism, surrealism, cubism, pop art. These styles range from those that resemble real world objects, leaning towards figurative art, to the other end of the spectrum using basic shapes and geometric deformations that diverge from reality.

% They also have a roughly anisotropic gaussians distributions for each of art modality, hinting biases to contours or high frequency data

\subsection{Background}
Computational art is a field that finds applications in the visual art industry for modeling collections of art \cite{ozgun2023computational}, simulating visual art for artwork exploration \cite{Stork2006}, and replicating artworks for preservation \cite{li2018dating}. \MV{As the field grows, it is essential to recognize that the distinction between human-created and AI-generated artwork remains subjective \cite{bellaiche2023humans, hertzmann2018can, AUGELLO201622}. The papers discussed in the literature use datasets from the art domain for training, thus modeling the domain, even if the art dataset is a subtask or a subset of the training dataset at least.
For clarity, we use the term 'AI-generated graphics' to refer to AI-generated works, acknowledging the ongoing debate over whether AI art qualifies as visual artwork. Additionally, when using the terms ‘art’, ‘visual art’ and ‘artwork’ we refer to the input data collection. The survey paper discusses research where AI image synthesis and generation models result in different interpretations of artworks from existing styles, diversifying visual arts datasets \cite{ernst2023artificial} or producing artistic counterparts to real-world images in the introduction and synthesis sections. AI tools assist in the extraction and analysis of growing collections of art to enhance interactive experiences while expanding interpretation across selected geometry priors in collections \cite{fan2023research, rani2023exploring}.}

It evolved from a form of static procedural art, building from preset rules with random perturbations through models such as the AARON model in the 1960s. Its later iterations were more data-centric, allowing for generation of more dynamic art using computer graphics through non-photorealistic rendering techniques that distilled artistic styles to simple parameters such as brush strokes and other learned statistics forming popular techniques such as style transfer \cite{hertzmann2018can}. These artworks were used to form interactive art pieces where both the artist and the audience influence the displayed design. Computational art evolved to include dynamic content generation with the evolution of image generative models, data-driven approaches learn how to synthesize AI-generated graphics through GAN-based models in the form of CAN, pix2pix, Cycle-GAN and GANVAS or learnable style transfer from DeepArt \cite{ anantrasirichai2022artificial}, digital art proliferated depictions of various media in different existing styles.
Other styles involved repurposing existing computer vision models for hallucinating emergent styles from natural images such as DeepDream. With a demand for controllable AI-generated graphics with the insertion and deletion of objects of varying poses and views, the AI art community developed text-to-image models such as DALL-E and Muse. Current iterations of image generative models aim towards high-quality AI-generated graphics using diffusion models such as GLIDE or Stable Diffusion. They currently suffer from mitigating data biases from the confounding of the style and color choices of particular art movements \cite{srinivasan2021biases} that result in hallucinating structures.
 
\begin{comment}
Art museums such as the Metropolitan Museum of Art, the Museum of Modern Art, and the National Gallery of Art use AI art in their exhibits \cite{rani2023exploring}. Others use computer vision for object detection and recognition, artwork cataloging and curation, 3D tours of sculptures, and augmented reality experiences with captions describing a piece. Some projects that use AI art include Duan et al.'s \cite{10.3389/fpsyg.2021.713545} emotion analysis to create personalized art derivatives based on Van Gogh’s paintings, Nawar's \cite{nawar2020collective} interactive art project exploring bread as a representation of one’s peculiar voice and political statement, Augello et al.'s \cite{AUGELLO201622} cognitive architectures for artificial agents creating paintings, and other deep learning tools used in creative processes and analysis of fine art. 
\end{comment}
% EH: reorganised and polished a bit
Art museums broadly use AI tools to collect user statistics regarding exhibit visitations and tours  \cite{rani2023exploring}. Additionally, they use computer vision in wider areas such as the Museum of Modern Art’s Thinking Machines exhibit use computational machines for artistic production\cite{james2018thinking}, or the National Gallery reconstructing old master paintings using imaging techniques developed in the Art-ICT conferences. The Metropolitan Museum of Art of New York City, colloquially referred to as ‘The MET’ even sources its data, thereby helping improve the retrieval and classification performance of these tools \cite{ypsilantis2021met}.
Others use computer vision for object detection and recognition, artwork cataloging and curation, 3D tours of sculptures, and augmented reality experiences with captions describing a piece. For example, Augello et al.'s \cite{AUGELLO201622} cognitive architectures for artificial agents creating paintings, and other deep learning tools used in creative processes and analysis of fine art. In addition to the more passive art exhibitions, AI art has been used for enhancing the interactivity in art, such as Duan et al.'s \cite{10.3389/fpsyg.2021.713545} emotion analysis to create personalized art derivatives based on Van Gogh’s paintings, and Nawar's \cite{nawar2020collective} interactive art project exploring bread as a representation of one’s peculiar voice and political statement.

%Art datasets are challenging for artificial intelligence techniques due to the generally smaller scales and larger inter-class variations \cite{redmon2016you, Mathieu2014}, as well as missing or inaccurate label annotations \cite{Milani2022, Elliot2013}. 

\subsection{Geometry in the Visual Arts Industry}
Identifying and analysis of artworks often use geometries ranging from global cues such as composition, perspective and proportions to identify stylistic characteristics common to an artist, to local cues such as stroke patterns, directions and shapes. Geometry is widely used in the art industry to represent perspective and lighting, as well as to reconstruct 3D shapes and locations of objects from 2D pictures. Models that incorporate proxy geometry onto artworks \cite{anantrasirichai2022artificial, Cox2015} find applications in animations and VR-/AR-based museum tourism. The learned proxy geometry are geometric features or model embeddings that learn style invariances or shape and geometric data information. The 3D proxy or 3D geometric features is an intermediate representation upon which these creative applications perform operations such as relighting and novel views from their 2D projections or 2D geometric features. The use of the generation of 3D generated models extends to content recovery for art conservation projects \cite{Pintus2016} in image searching for art historians and experts. Depending on the art style the object geometry is similar or exaggerated compared to their real-world counterparts \cite{cohen2022semantic}. The geometry data, such as pose \cite{PRATHMESH2022}, keypoints  \cite{lorente2021museum} or bounding-box \cite{Ufer2021}, can then be used as labels to retrieve and match images with objects that range from highly structured to highly varied geometries. They also take the form of extra input maps along with the photographs of artworks like murals or paintings on surfaces like pots, walls and robes \cite{borg2020application} provide extra information when projected together to form 3D models. Table 1 shows examples of visual arts datasets using such geometric data for various tasks discussed over the duration of the survey paper.
A 3D proxy is an intermediate representation upon which these creative applications perform operations such as relighting and novel views from their 2D projections. The use of the generation of 3D models extends to content recovery for art conservation projects \cite{Pintus2016} in image searching for art historians and experts. Depending on the art style the object geometry is similar or exaggerated compared to their real-world counterparts \cite{cohen2022semantic}. %This labels such as pose \cite{PRATHMESH2022}, keypoints \cite{lorente2021museum} or bounding-box \cite{Ufer2021} to retrieve and match images with objects that range from highly structured to highly varied geometries. 
The geometry features, such as pose \cite{PRATHMESH2022}, keypoints \cite{lorente2021museum} or bounding-box \cite{Ufer2021}, can then be used as labels to retrieve and match images with objects that range from highly structured to highly varied geometries. 
Extra input maps along with the photographs of artworks like murals or paintings on surfaces like pots, walls and robes \cite{borg2020application} provide extra information when projected together to form 3D models.

Computer vision and machine learning for extraction and analysis from art collection meta data and images or \MV{3D} models provide an alternative to visual formal analysis of art collections that can be subjective, time intensive and inconsistent between experts. Geometric priors from data or those learned from models help detections in cases such as complex or cluttered scenes \cite{Milani2022, Ufer2021, Kadish2021}, exaggerated or abstract poses \cite{Jenicek2019, Pramook2016,Qingfu2020, Islam2011, Marsocci2021}, and perspective distortions or low dynamic range \cite{bernasconi2023computational, 10.1145/3490100.3516470}.
 
Additionally, variants of such geometric conditionals find use in generative art to create controlled and diverse outputs while preventing repetitive or biased patterns that may emerge from the underlying data or model generation process \cite{soddugenerative}. This mitigates problems such as incoherence \cite{farid2022perspective}, stereotypes and prejudice \cite{luccioni2023stable} in current AI-generated graphics that lead to texture bleeding from parts of objects to each other, deformed hands or badly constructed objects. With aspects of geometry modeled separately in the model through proxy objects, shading and illumination stages, researchers can even model impossible, inconsistent and incoherent shapes in input painting images \cite{akleman2024hyper} on purpose.

% The inclusion of geometric information at the data preprocessing stage and as labels induces model invariances during training. 

\subsection{Paper Organization}

 To understand how geometry contributes to artwork tasks, we discuss the artificial intelligence techniques facilitating the use of geometry in extracting, analyzing and synthesizing artworks. We believe that discussing extraction, analysis, and synthesis is pertinent as these are the primary applications of AI tools in the art industry, utilized by experts, critics, and visitors in art collections. These processes are interconnected through the common thread of geometric considerations in various learned representations or additional constraints, which play a crucial role in enhancing the understanding and appreciation of art. We aim to emphasize the evolving nature of art and how AI tools are increasingly being integrated into creative processes and their interpretations. The focus on geometry in this discussion aims to improve the quality of the generated media by providing form guidance amidst the fluidity and variety of styles while allowing more control for artistic expression. Additionally, it addresses common failings in these AI tools \cite{Jenicek2019} with regards to generated images, visual composition and geometric deformations.
 
We first discuss the extraction of geometric labels for humans and objects from 2D images to 3D models in \textbf{Section \ref{section:extraction}}. The object labels are divided into bounding boxes, key points and segmentation masks whereas people range from pose skeletons, landmarks and gestures. The 3D features range from explicit surfaces to implicit surfaces, and parametric models. Then, we explain the analysis of the effectiveness of the extracted geometric data on discriminative tasks in \textbf{Section \ref{section:analysis}}. The feature extraction section discusses the extraction of entities or subsets of visual art collections where geometric information is used directly as constraints or selection criteria to improve discrimination or indirectly by augmenting the collection to improve model classification.  Next, we detail the synthesis and manipulation of artwork for novel view synthesis, relighting and content restoration in \textbf{Section \ref{section:synthesis}}. The section mainly discusses the use of geometry for visual art collection modification which provides new perspectives or renditions of existing artwork where any changes made are geometrically and stylistically consistent with the original artist's vision. Additionally, it discusses content recovery where the geometric consistency is towards the original artistic medium in addition to the entity's geometry for both global and local consistency preservation. Finally, we discuss the \textbf{future directions} for better incorporating geometry into the model architecture such that it is fully differentiable to use the full strengths of deep learning methods.

\subsection{Related Surveys}
%Geometric information in artificial intelligence gives representations that encode the inherent structure of visual elements. We 
Using Geometric information in artificial intelligence models facilitates the learning of representations that encode the inherent structure of visual elements. This paper covers 2D and 3D artistic visual media while discussing the extractable geometric features following it with the discriminative and generative tasks they can be incorporated into. The closest work related to ours considers geometric features at the local and image level through feature descriptors or hardware (e.g. 3D printers and scanners) for 3D models in cultural heritage \cite{pintus2016survey}. Unlike our work, they focus on preservation, registration, reconstruction and enhancement, and do not consider deep learning based techniques. In this section, we first cover more recent surveys in the field before covering the works that incorporate AI and geometry in visual art.

AI-based methods have found use in creative applications such as content generation in multimedia, captioning, spoofing and AR/VR \cite{anantrasirichai2022artificial}. These involve models deployed for production in games such as GameGAN, storyline generation with MADE or Vid2Vid, and artwork generation from Hypercube-based NEAT that utilizes geometric regularities. A survey on computer vision in art history highlighted its applications in image search and retrieval \cite{foka2021computer} for art historians. They identify the importance of recognizing contexts such as clothing, architecture, materials, faces, patterns on objects, and artist signatures for recognizing time periods, geography and culture. Other papers focus on deep learning approaches in paintings \cite{castellano2021deep} and digital art collections \cite{cetinic2022understanding} for content recognition such as classification, retrieval and detection in images or multimodal domains. They also cover a subset of art synthesis using image generative models with losses that exaggerate styles or through latent space guidance with other models such as Contrastive Language-Image Pre-Training (CLIP). We cover the discriminative and generative tasks only if they consider geometric information in the form of annotations or pseudo geometry in the form of intermediate representations, model functions or dataset transformations to induce model invariance.

A study on mixed 2D and 3D non-photorealistic media covers the task of art conservation using multiple input spectrums \cite{borg2020application}. They only use machine learning or statistical information for detection when dealing with the visual spectrum, with the majority of their study covering technologies for diagnosis and imaging. A similar review covers these multispectral inputs for paintings only, which extracts 3D geometric information through correspondence matching with feature descriptors such as SIFT \cite{remondino2011review}. However, we explore not only art restoration but also the repurposing of material properties, symmetries, and corrupted regions within 3D representations. This enables us to simulate missing contents lost due to deterioration or the image sensing process.

% The green boxes outline the cluster centres with K-means and K=10 selected as the clustering algorithm.

\begin{sidewaystable}[]
\begin{tabular}{|l|l|l|l|l|}
\hline
Source                                                                     & Dataset                                                                             & Task   Description                          & Dataset   size & Number   of classes \\ \hline
\cite{castellano2021visual,   9412438}                    & best-artworks-of-all-time   (Kaggle dataset)                                        & Image   retrieval                           & 8.4K           & 50                  \\
\cite{Madhu2022}                                          & Christian   archeology (CHA)                                                        & Object   detection                          & 16K            & 16                  \\
\cite{ahmad2023toward}                                    & StyleObject7K                                                                       & Object   detection                          & 7K             & 10                  \\
\cite{ahmad2023toward}                                    & ClipArt1K                                                                           & Object   detection                          & 1K             & 8                   \\
\cite{ahmad2023toward}                                    & Watercolor2K                                                                        & Object   detection                          & 17.8K          & 6                   \\
\cite{ahmad2023toward}                                    & Comic2K                                                                             & Object   detection                          & 52.7K          & 6                   \\
\cite{Fuertes2022}                                        & QMUL-OpenLogo                                                                       & Logo   detection                            & 27K            & 352                 \\
\cite{Kadish2021}                                         & StyleCOCO                                                                           & Human   detection                           & 61K            & -                   \\
\cite{smirnov2018deep,zhao2023research,   saleh2015large} & WikiArt   Paintings                                                                 & Object   detection                          & 81K            & 27                  \\
\cite{Jeon2020}                                           & Brueghel                                                                            & Object   detection                          & 1.5K           & 10                  \\
\cite{kadish2021improving}                                & People-Art                                                                          & Human   detection                           & 4.5K           & -                   \\
\cite{smirnov2018deep}                                    & IconArt-v2 (IA).                                                                    & Object   detection                          & 6.5K           & 10                  \\
\cite{zhao2023research}                                   & Artsy   scrapped dataset                                                            & Orientation   detection                     & 2.8K           & -                   \\
\cite{sandoval2021adversarial}                            & Pandora   18K                                                                       & Style   classification                      & 18K            & 18                  \\
\cite{anwer2016combining}                                 & Painting-91                                                                         & Style   classification                      & 4.2K           & 13                  \\
\cite{castrejon2016learning}                              & CMPlaces                                                                            & Scene   classification                      & 2.5M           & 205                 \\
\cite{cohen2022semantic}                                  & DRAM                                                                                & Semantic   segmentation                     & 2.5K           & 12                  \\
\cite{huang2023controllable}                              & ArtSem                                                                              & Semantic   segmenation and Image generation & 40K            & 5                   \\
\cite{Wechsler2019}                                       & MAFD-150                                                                            & Face   detection                            & 150            & 29                  \\
\cite{PRATHMESH2022}                                      & ClassArch                                                                           & Human   pose estimation                     & 1.7k           & -                   \\
\cite{Yaniv2019}                                          & Artistic-Faces                                                                      & Face   detection                            & 160            & -                   \\
\cite{Pramook2016}                                        & Drawings                                                                            & Human   pose estimation                     & 2.5K           & -                   \\
\cite{STEFANIE2023}                                       & Poses   of People in Art                                                            & Human   pose estimation                     & 2.4K           & 22                  \\
\cite{ciortan2021colour}                                  & Dunhuang                                                                            & Image   Inpainting                          & 5.6K           & -                   \\
\cite{xue2021end}                                         & \begin{tabular}[c]{@{}l@{}}Chinese-Landscape-\\    \\ Painting-Dataset\end{tabular} & Conditional   Image generation              & 2.1K           & - \\ \hline       
\end{tabular}
\caption{Artwork datasets used in the training of models discussed in the Geometric Features Extraction, Discriminative Geometric Features Analysis, and Synthesis with Geometric Features sections. The table describes the dataset names, the tasks they are utilized in along with the total size of the dataset for training, testing and validation. Finally, the number of classes in the dataset is counted, with entries left blank for single-class datasets or those using paired inputs and outputs, such as images with their ground truth semantic maps or pose skeletons.}
\label{tab:datasets}
\end{sidewaystable}

\section{Geometric Features Extraction}
\label{section:extraction}

% Difficult fusing things together in the feature extractor section
% \usepackage{multirow}

\begin{sidewaystable}
\begin{tabular}{|llllll|}
\hline
\multirow{2}{*}{Geometric data}           & \multicolumn{5}{c|}{2D Representation}                                                                                                                                                                                                                                                                                                                                                                                                                                                                                                                                                                                                                                                                                                                                                                                                                                                                                                                                    \\  
                                          & \multicolumn{1}{l}{Bounding box}                                                                                                                     & \multicolumn{1}{l}{Keypoints}                                                                                & \multicolumn{1}{l}{Object segmentation}                                                                                                & \multicolumn{1}{l}{Landmarks}                                                                                                                                         & Pose skeleton                                                                                                                                                                                                                                                                            \\ \hline
Existing work & \multicolumn{1}{l}{\begin{tabular}[c]{@{}l@{}}\cite{Jeon2020, Ufer2021, Milani2022, kadish2021improving, Bai2021, shen2019discovering, cai2015cross, Jeon2020, Fuertes2022, Joseph2011, gonthier2018weakly, Elliot2013}\\ \cite{Lu2022, Marinescu2020,Sizyakin2020, 1467360, Nicholas2016, Elliot2014, Lang2018, Madhu2022,Wechsler2019} \end{tabular}}                                                                    
& \multicolumn{1}{l}{\begin{tabular}[c]{@{}l@{}}\cite{Sizyakin2020,Elliot2014, STEFANIE2023,delgado2023crossing}\\ \\ \\ \end{tabular}}                            
& \multicolumn{1}{l}{\begin{tabular}[c]{@{}l@{}}\cite{Milani2022,Marinescu2020, doi:10.2352/EI.2022.34.13.CVAA-169, cohen2022semantic}\\ \\ \\ \end{tabular}}                                                      
& \multicolumn{1}{l}{\begin{tabular}[c]{@{}l@{}}\cite{Zhang2023,Matthias2022,Yaniv2019,Aline2022, Zhang2023}\\ \\ \\ \end{tabular}}                                                                                     & \begin{tabular}[c]{@{}l@{}}\cite{Jenicek2019, Pramook2016,Marsocci2021,1467360, Islam2011, Qingfu2020} \\ \cite{Carneiro2012,Matthias2022,ju2023humanart}\\ \\ 

\end{tabular}

\\ 
\hline

Techniques used                           & \multicolumn{1}{l}{\begin{tabular}[c]{@{}l@{}}Single stage detectors\\ Multi stage detectors\\ Transfer learning\\ Multimodal learning\end{tabular}} & \multicolumn{1}{l}{\begin{tabular}[c]{@{}l@{}}Correspondence \\matching\\ Point-source detection\end{tabular}} & \multicolumn{1}{l}{\begin{tabular}[c]{@{}l@{}}Segmentation models\\ Conditional maps\\ Saliency maps\end{tabular}} & \multicolumn{1}{l}{\begin{tabular}[c]{@{}l@{}}Off-the-shelf face \\ detection\\ Landmark \\ transformation\\ Geometric style transfer\\ Region networks\end{tabular}} & \begin{tabular}[c]{@{}l@{}}Transfer learning on\\ pose estimators\\ Multimodal pre-training\\ Pose matching \\ and clustering\end{tabular}   \\ \hline
Visualization                             
& \multicolumn{1}{l}{
\begin{minipage}{.12\textwidth}
      \includegraphics[width=1\linewidth, height =1.3\linewidth]{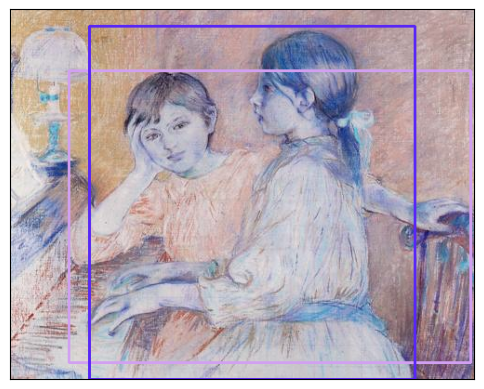}
\end{minipage}  
}                                                                                                                               
& \multicolumn{1}{l}{
\begin{minipage}{.12\textwidth}
      \includegraphics[width=\linewidth]{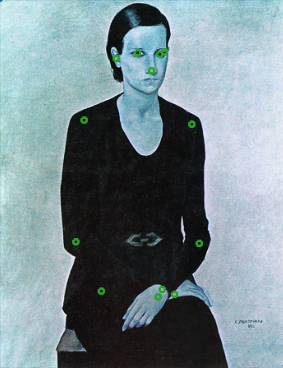}
\end{minipage}
}                                                                                         
& \multicolumn{1}{l}{
\begin{minipage}{.12\textwidth}
      \includegraphics[width=\linewidth]{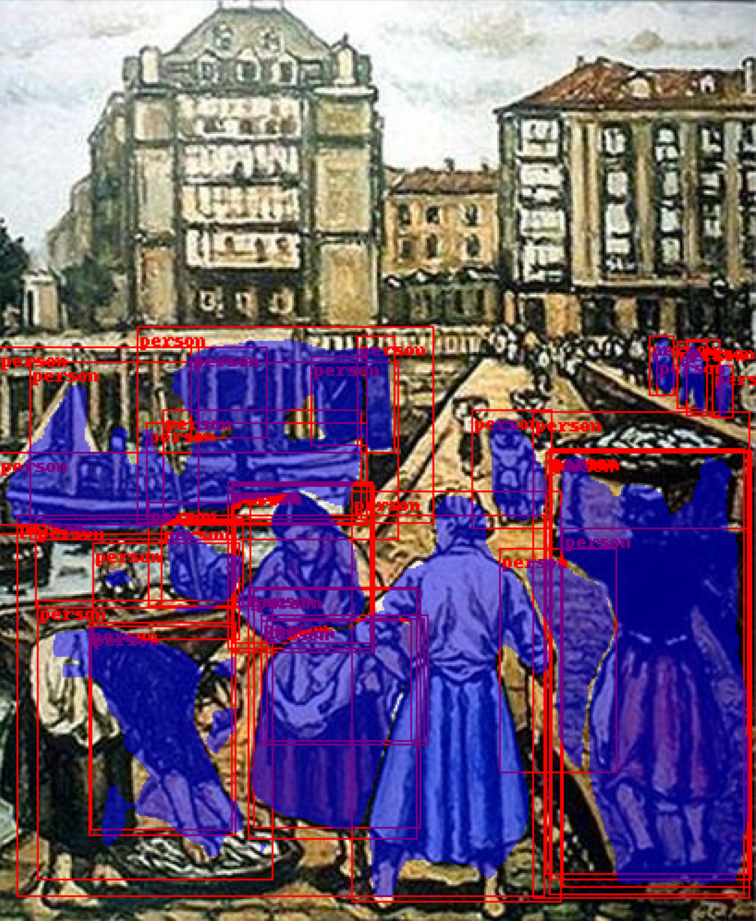}
\end{minipage}  
}                                                                                                                 
& \multicolumn{1}{l}{
\begin{minipage}{.12\textwidth}
      \includegraphics[width=\linewidth, , height= 1.1\linewidth]{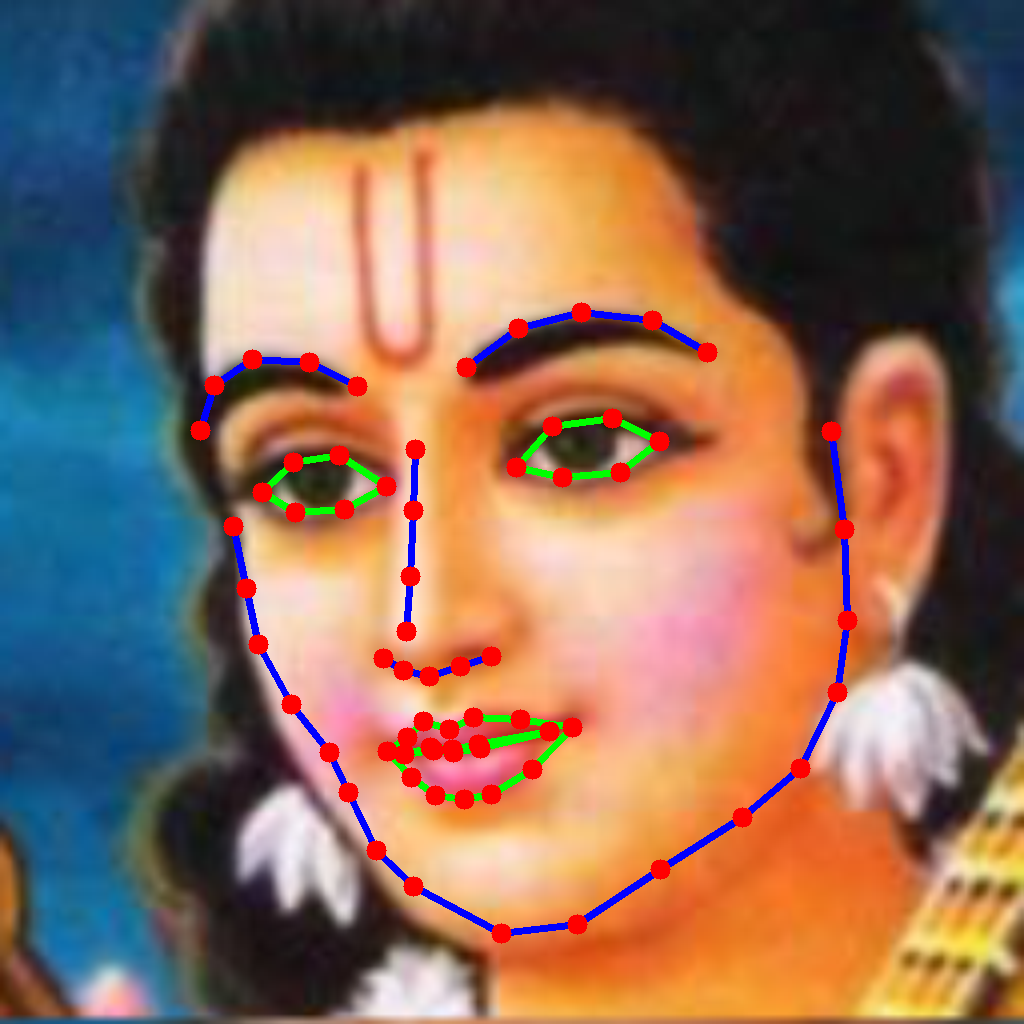}
\end{minipage} 
}                                                                                                                                                 
&                                                                                                                        \begin{minipage}{.15\textwidth}
      \includegraphics[width=\linewidth]{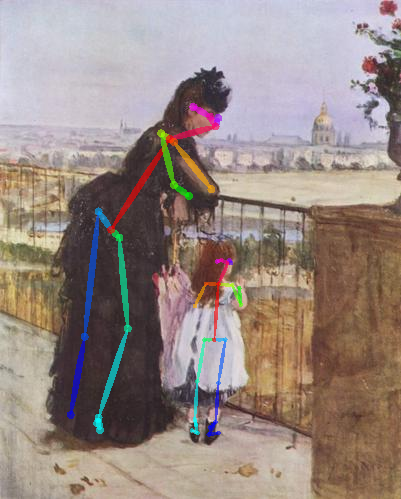}
\end{minipage}

\\ \hline
\end{tabular}
\caption{Different types of geometric labels extracted for objects and humans ranging from bounding boxes indicating scale and position, points of interest indicating correspondences, maps and masks for important regions, and landmarks and pose skeletons indicating structure. The table lists the techniques used to extract these geometric labels in the Extraction section for images of objects and humans. The bounding boxes better localize people in backgrounds with higher contrast for the impressionist painting of two children sitting with a piano. The keypoint detection here detects pose as the points of interest in a portrait image of a lady. The segmentation task provides shape information through the mask along with a bounding box to enclose the extent of the boundary for the people on the pier and the boats. In landmark detection, the structure of the face of a Hindu God is annotated with points and the contour connecting them.}
\label{table:extraction_2d}
\end{sidewaystable}

Object detection uses geometric features that act as descriptors learned from special classifier architectures to learn external geometry data \cite{smirnov2018deep} from labeled boxes or pixels that enclose the object using bounding boxes or semantic maps. These descriptors can be specialized \cite{nguyen2016human} for human detection by adding more structure and robustness to affine transformations via pose skeletons at different hierarchical levels. Common geometric data capture regions of interest for both humans and objects at the semantic and instance level. The latter has not been explored in the existing works in favor of the problems of domain adaptation and limited data. Some data augmentation techniques and extra input induce style invariance and other context invariances such as time periods respectively to force models to process the input data as geometric feature embeddings that capture structure information, thereby forming geometric techniques.

Paintings are challenging to computer vision models due to their background clutter and object composition \cite{Hall2015}. %Due to 
With their high diversity in poses and shapes as compared to their real-world counterparts, they can even lead to spurious detections, mistaking people for other mammals \cite{Nicholas2016} or from occlusions. Furthermore, some painting datasets do not have reliable annotations with some subjects missing. To relax the problem, existing works account for the deviation of the geometry and depictions from the real world to \MV{the artistic image domain} with modifications to their real-world detector training and inference pipelines.

\subsection{Object-Centric Features}

Object-based geometrical feature annotations involve varying amounts of information contained in excess for bounding boxes, exactly in segmentation masks and minimally in keypoints. Alternatively, data pre-processing induces the underlying class of transformations during model training or removes non-geometric information. These labels are summarized and visualized in Table \ref{table:extraction_2d}.

\subsubsection{Bounding Boxes}
Bounding boxes provide the target objects' position and scale in paintings with rectangles by manual or automatic annotation using object detection techniques. Object detection is divided into single-stage or two-stage models for speed-accuracy trade-off, and scores the overlap between their proposed regions of interests and the bounding boxes \cite{Ershat2022}. In the painting domain, these models account for the gap between the artistic depiction of the object and its real-world counterpart by transforming the input data \cite{Jeon2020} or modifying its stages \cite{Ufer2021, Milani2022, kadish2021improving, Bai2021} depending on the architecture choice. Modifications to one-stage models proceed in an end-to-end fashion to classify detected objects while multi-stage detectors optimize their constituent stages to produce better candidate regions \cite{shen2019discovering}.

%\textbf{Paragraph header:} 
Single-stage detectors such as You Only Look Once (YOLO) directly classify objects and perform regression using Convolutional Neural Networks (CNN) to get their location and size from the predicted box coordinates and object aspect ratio. They utilize spatial consistency in datasets to implicitly embed the geometric and content information together and do not use intermediate features to represent object regions. In \MV{fine-art paintings}, associations between objects provide a non-destructive means for identifying visual connections for investigations into its history or authenticity \cite{shen2019discovering}. However, datasets with nonnatural collections lack strong spatial correspondences among highly diverse objects, limiting the effectiveness of pre-trained detectors. Furthermore, the shape abstractions in paintings and sketches vary drastically from photographs in the resultant high-level feature space \cite{cai2015cross}. To address these challenges, detectors often employ data augmentation techniques like style transfer \cite{Jeon2020, Fuertes2022} to align real-world images with \MV{artistic image} styles and structures, thereby increasing the \MV{artistic images} dataset size. However, these augmentation techniques require further processing to maintain semantic consistency \cite{ernst2023artificial} since failed samples actively detriment model performance. These augmentations may compromise semantic consistency \cite{Jeon2020} unless complemented by techniques such as mask content mixing, which combines valid regions of objects from multiple images without overlap or obfuscation \cite{Fuertes2022}. Nevertheless, style transfer is useful for texture-biased models like CNN-based object detectors \cite{Jeon2020}, unlike shape-biased models \cite{10.3389/fpsyg.2021.713545}. For example, style transfer is unsuitable for certain \MV{artistic image styles} such as stick figures and sketches \cite{kadish2021improving} where careful content mixing that preserves semantic correspondences is preferred instead \cite{wang2023cut}.

Multi-stage detectors add an additional feature extraction stage before classification and regression which customizes the learned object representation, size and shape. It is commonly based on Region-Based Convolutional Neural Networks (R-CNNs) that have a feature extractor to extract and warp the regions of interest into a uniform aspect ratio with a CNN for model prediction \cite{Ershat2022}. In paintings, the learned deep features help detect near duplicate objects that only differ in style \cite{shen2019discovering} that pay homage to popular artists and schools by copying or modifying their composition. This ability enables tasks, like grouping styles or achieving texture, invariance by replacing \cite{Ufer2021} or modifying architecture stages \cite{Milani2022}. Earlier methods extract static features using style templates \cite{Ufer2021, Joseph2011} to capture the appearance variation of an object so that the feature extractor works with a robust representation that accounts for a small variation of style, color, scale and orientation. The template-based detection finds use in objects with standard configurations such as \MV{paintings} authentication which requires artist signature attribution from small motifs in architecture and heterogeneous paintings of still life, portraits and landscapes. Some later work improves parts of pre-trained R-CNNs with multi-head attention modules \cite{Milani2022} for improved model performance to extract and exaggerate the most class discriminative subregions to improve detection for cases in the Art-DL dataset's rarer classes Dominic and Paul or those that co-occur such as Mary and Jesus. The selection stage in detectors helps separate the foreground from the background in crowded scenes composed of objects of varying sizes, unlike single-stage detectors. Furthermore, unlike single-stage detectors, the modifications to the feature extractor allow generalization to object geometry in terms of scale and proportions, but they limit the adaptability to novel tasks based on the choice of the feature extraction strategy. Depending on the method of feature representation learning, detectors can enhance their performance for classes with small sample sizes, which are often considered anomalous in accordance with their training dataset. This process involves separating the inputs' appearances from their other attributes \cite{liu2023generalized, liu2023amp}. In cases where the classes are not imbalanced, the representations can be robust enough to handle label shifts or produce consistent activations with specific groups  \cite{liu2023learning, liu2023distributional}. 

Transfer learning involves training the detector's pre-trained layers on a different data domain, using additional components and data augmentation to adapt the model for a new task. The performance of a pre-trained model relies on how the second stage of multistage detectors is trained. These training choices may involve training parts of the models in stages to bias the detector toward the desired task. Alternatively, bootstrapping the model can provide weak supervision by using a small subset of clean, labeled images to propagate to the remaining data \cite{gonthier2018weakly}. For Japanese Ukiyo-e paintings, although pre-trained models accurately detect faces, the model's classification of the cropped face bounding boxes is poor without fully tuning the model weights \cite{vijendran23tackling, feng2021rethinking}. Fully fine-tuning classifiers can significantly improve model performance, especially when coupled with data-efficient learning methods like contrastive learning \cite{feng2021rethinking}. However, this approach depends on the sampling strategy for the training data, and the compatibility of data augmentation with the pre-training dataset and task at hand.

Instead of retraining or modifying a model for a specific task, we can leverage information from multiple domains and utilize relationships in the latent space between multiple models for domain adaptation with multi-stage detectors. For instance, multitopic language modeling \cite{Bai2021} can provide context for image captioning by using scene relationship embeddings from the \MV{painting} description dataset SemArt alongside pre-trained Faster R-CNN and Residual Network (ResNet) models. This approach requires fewer annotations for generating \MV{image} descriptions, but the domain gap among the models results in significant variation in image description evaluation metrics. In some image captioning tasks \cite{Lu2022, Bai2021}, the object detector acts as either a feature extractor or a caption generator. Features from different modalities' outputs are fused using the attention mechanism of transformers. The detectors play dual roles as image captioners and intermediate feature extractors, facilitating semantic alignment at either the image-text \cite{Bai2021} or image-image \cite{Lu2022} level, thereby bridging different domains. In the former case, the detector can also categorize detected objects into a hierarchy of textual categories. Moreover, semantic metadata can enhance object detection performance \cite{Marinescu2020} by filtering out objects incompatible with the periods depicted in scenes, eliminating the need for auxiliary models.

\subsubsection{Patch-Based Region Selection}

In the absence of labelled bounding boxes, patch-based selection methods approximate object locations with feature engineering and simple classifiers. They enforce faithful geometry through spatial correspondences \cite{shen2019discovering} embedded in model features with a further selection stage that accounts for a particular class of transformations. These matches at the feature level narrow down the area of interest from the image to a patch level to return the target object. Object retrieval for paintings typically uses a multistage model for feature engineering and better detection and localization using part-based models. These multistage detectors are useful for noisy datasets where the target classes are absent in the object. The first stage clusters related image-level candidates through text mining and choosing patch-level candidates through segmentation and MLDPs through a Histogram of Gradients (HoG) while the second stage trains class-wise DPMs as a detector. 

During the feature engineering stage, the fusion of different data modalities provides the model with a shared representative space within the same domain \cite{Sizyakin2020} or between different domains \cite{Bai2021} and robustness to noise from different sources. Additionally, their feature embeddings often require heavy pre-processing with dimensionality reduction into embeddings such as fisher vector\cite{ Elliot2014}, histograms and templates to encode image content. Other common embeddings involve model-specific image embeddings such as HoG as a learned feature encoding method \cite{1467360} for a middle-level discriminative patch as a robust template matching alternative after hyperparameter tuning. 

For detection, part-based models work with a smaller number of proposals and utilize the patch statistics alongside coarse geometric information from the preprocessed features. Deformable Part Model (DPM) \cite{1467360} is a popular choice as a class-specific sliding window detector \cite{Elliot2013, Nicholas2016} since it is robust to object detection under a lot of arrangements of its sub-parts. Detector robustness is crucial for domain adaptation in problems such as detecting gods or animals in vases \cite{Elliot2013} where there is a lot of subject ambiguity. Siamese networks as the detector architecture provide another solution by contextualizing the data as a strategy to capture the scene with co-occurring object embeddings \cite{Madhu2022}.

\subsubsection{Geometric Transformations for Content Selection}
Simple geometric transformations like affine transformations and cropping \cite{Smirnov2018} also help models train on small datasets while learning invariance properties from the data. %providing invariance to their underlying properties.

Data augmentation techniques provide the additional benefit by mitigating the need for extensive data labeling by generating additional training examples that capture variations in object appearance and their contexts. There are techniques for generating new data from the image level to preserve the training distribution. LogoMix \cite{Fuertes2022} \MV{synthesizes samples from overlapping different logo patches, effectively preventing the model from overfitting to synthetic data and ensuring it learns robust features from real-world data.} % disallowing the model to overfit to the synthetic data instead of the training data.

%style transfer - weak, detection improvements
Style transfer creates a hybrid image that preserves the structure of the content image of the real-world while its style takes after the painting images. Other data augmentation techniques often drastically transform the training distribution to account for the gap in diversity between domains \cite{Da2017}. They can have fidelity towards different aspects of the input pair depending on if they were trained on multiple styles, a single style, or through iterative optimization. The synthetic data aggressively transforms the training dataset, provided the input data is not independent of texture such as edge maps while needing only a limited set of style images\MV{.} 

Style transfer also suffers in diversifying images with structures \cite{Kadish2021} like stick figures, but it can still provide robustness to textural distortions to adapt object detection in the painting domain. Their classifier uses ResNet-152 and Faster-RCNN that are fine-tuned on stylized COCO dataset and showcases the improvement in detection on larger training datasets, even if they are simply stylized versions. While style transfer provides a large shift in the color and texture information, without trading off the content loss fidelity towards the style image, it does not correspondingly warp shapes. 
In object detection, style transfer helps reduce the cross depiction problem to that of color and structure \cite{Lang2018} attributes in the image while ignoring geometric information that is context dependent like gesture, shape and pose. Artistic representations can be more shape-biased \cite{Kadish2021} for stronger geometric features and better model performance. The Computer Vision Group (CVG) system \cite{Lang2018} performs image retrieval using contours from images for stroke information and negative example training after expert annotation with five bounding boxes to determine spatial extent and geometric relation.

\subsection{Human-Centric Features}
Human detection on the other hand involves highly regular structures at the face, body and hand level captured by bounding boxes, pose skeletons and landmarks. 
Pose and body shape information provides information for perception and identities \cite{Islam2011} through the proportion of their parts. Human poses can vary in depiction across time periods and represented with different topology or motifs \cite{STEFANIE2023} where the referential deviations can represent artistic signatures or movements. They can also increase the difficulty in detecting poses by blurring contours, distorting proportions or occluding joints through apparel or other objects or lighting, sometimes even changing the cardinality of the parts to indicate mythological creatures. Face detection in uncontrolled settings like modern \MV{artistic styles} \cite{Wechsler2019} is very challenging due to occlusion as well as variations in shape, color, texture and face size. In end-to-end training, the ratio of the preservation of the geometric features or the pose regression loss to the style transfer loss is vital, unlike training them as separate parts for achieving high pose accuracy \cite{Qingfu2020}. 
There is still a cross-domain generalization problem since the joint positioning is more accurate with real-world images compared to artistic images, but more data from stylization gives better performance after a cutoff. Paintings retrieval can benefit from pose annotations at different levels of abstractions \cite{Carneiro2012} to compensate for mislabeling. %categories. 
Inverted label propagation produces these levels of poses through producing annotations induced from the source to the target image provided that the dataset size is sufficiently large. 

\subsubsection{Hand Gestures}
The detection of hand poses involves the position and orientation of the hands and their fingers with respect to the body in the form of templates or pose skeletons. In portraits and paintings, they commonly form hand signs and iconographic meaning with irregular finger positions or unnatural gestures with hand actions \cite{lazzeri2019secret}. They indicated a group's, family or religious memberships and ranks, personality traits and an artist's signature style. Learnable hand templates \cite{Joseph2011} strongly separated the hand from the background, which convolve with Laplacian of Gaussian filters across the image and give strong responses on contour alignment. While their collections capture the primary variation of appearances across scale and rotations with data augmentations and principle component projections, they do not encode relations between the hand and body. This results in detecting false positives for speaking gestures and other semantically ambiguous actions. 

Follow-up works found that the use of deep pose estimators alone results in poorer accuracies in \MV{western fine-art paintings} \cite{bernasconi2023computational, 10.1145/3490100.3516470} due to distortions of perspective and low contrast of the body against the background. They used the OpenPose model \cite{zheng2023deep}, a multistage convolutional model to detect and match part-wise confidence and affinity maps, to detect the skeletal pose of the hand with 21 keypoints and if it is left or right-handed. The model learns body part relations, their locations and orientations with their corresponding confidences with learned heatmaps and vector maps. The detection improves with better representations learned by pose descriptors. For example, the ResNet-50 pre-finetuned on a large sign language dataset can effectively recognizes gestures involving both hands when compared with their simple angle pose key point descriptor \cite{bernasconi2023computational}, but fails on the less represented classes with hand-object interactions. %hands with objects.  
Such misclassifications also result from low interclass variations in gestures in which similar poses belong to different classes.

%Such misclassifications \HS{do not use two result in the same sentence} also result from low interclass variations in gestures result in similar poses belonging to different categories.

\subsubsection{Facial Landmarks}
Facial landmarks in paintings have more variations compared to their real-world counterparts leading to its architectures disentangling the style and pose into separate multistage models or a need for data augmentation.

Multistage models such as the Cascaded Pose Transform network (CPTNetV2) can simulate head and face pose animation \cite{Zhang2023} to model pose displacements while inpainting the facial features separately to disentangle the problem into the two separate poses transformations. These models need a refinement stage to add details while maintaining consistency. Then, a fusion generator utilizes both the pose information that was disentangled by imposing masks to guide their individual generations. Other two-stage models can detect modalities like bounding boxes and keypoints for the human figure \cite{Matthias2022} from the photograph domain to that of paintings using a semi-supervised learning method with transformers through a teacher-student model distillation. It predicts a fixed set of proposals for each image, removing the need to account for overlapping boxes and imbalance between the foreground and background. Distilling geometric information for domain adaptation provides better results than fine-tuning or style transfer with additional label conditioning.

Artistic augmentation\cite{Yaniv2019} for landmark detection requires image transformation through style transfer techniques followed by a part-based feature correction step for landmark warping to account for structural shifts and decorrelating parts. Techniques like part-based correction and tuning and Geometric style transfer account for extreme styles and higher variation in landmark points. The stylized portraits' landmarks are warped to the mean facial shape vector of the target style to capture a signature structure using Thin Plate Spine (TPS) interpolation, but the method cannot handle fundamental shape variations from natural faces in anime or manga portrayals. A ResNet encoder-decoder and region networks can account for global and local landmark arrangements \cite{Aline2022} and result in accurate prediction of inner facial features in high-resolution images. Style transfer and geometric augmentations, to randomly shift or resize facial landmarks along with their movements on a TPS displacement field, account for their changed arrangements in artistic faces. Despite suffering from jawline landmark localization due to ambiguous labeling, the method works well for salient portrait features for the eyes, nose and mouth.

\subsubsection{Body Skeleton}
%pose similarity, body - weird, this one would be better at the start
Poses between people in paintings and photographs \cite{Jenicek2019} can be effectively aligned by pose detections and matching them through geometric transformations. The latter validates and measures similarity under different scales and positioning while making the detection robust to noise and missing parts. The method is not robust to unknown poses, occlusions, ambiguous poses,  or any spurious connections that arise from these challenges. Style transfer can bridge the gap between photographs and paintings for both person and pose detection in curvilinear surfaces like vases and create a dataset to fine-tune the HRNet~\cite{9052469} model for the tasks. With a perceptual loss on both tasks, the model can adapt the annotations to the pose and detection losses with the stylized data.

To compensate for limited painting data, the pose estimators can also be pre-trained from 3D renderings \cite{Pramook2016} of artistic media like anime or manga which provides joint positions from the underlying rigging. These models can simply be fine-tuned on a smaller dataset of drawings to effectively ignore the problem of domain gap from models pre-trained on photographs. The pose information can be utilized in other tasks such as image retrieval. Pose similarity followed by clustering helps retrieve similar paintings \cite{Marsocci2021} through methods like K-medians with metrics which are invariant to scaling, rotations and translations after detecting them through pre-trained models like OpenPose. 

\subsection{Segmentation Masks}
Image segmentation partitions the image into pixels that group into multiple classes which can be further grouped into individual objects that belong to the same class in the case of instance segmentation. This requires fine localization of objects in the scene regardless of scale, occlusions from clutter or other objects, or appearance changes from lighting or environmental conditions. Earlier works use deformable models, which provide an object shape template representing a distribution of warped objects, and graph cut to partition an image into regions while providing boundary separation \cite{he2011image}. By rephrasing the image segmentation as a deformable model optimization problem, they represent the Chinese paintings by their unique color choice in neighborhoods and the direction of texture. The deformation model splits the image into connected regions, while the texture directions represent flexible sparse foreground/background features like edge convolution filters to detect orientations.

Deep learning based segmentation networks like DeepLab v3, a fully convolutional model segmenting objects at various scales with spatial pyramid modules and cascading dilated and upsampling convolutions, transfer well to the artistic domain with transfer learning and style transfer \cite{doi:10.2352/EI.2022.34.13.CVAA-169}. When trained only on natural images, the baseline model on modern human portraits outputs faulty segmentations due to weak lighting cues, different color and texture choices compared to photorealistic images, and similar contours in the object and its background (e.g. striped sleeves and sofa). %They get better results when fine-tuning 
The model improves when fine-tuning once on style-transferred images, before training it on the real portrait images. Despite the unnatural color tone transfer in some regions, the model only fails with extreme shadows, reduced style, flat regions and unnatural skin tones. 

Similarly, style transfer augmentation \cite{kamann2020increasing, cohen2022semantic} improves the result of segmentation models by increasing their shape-bias while simultaneously providing robustness against various image corruptions (e.g. noise, blur, adverse weather conditions such as fog, motion blur from fast-moving objects). By separating the task into coarse binary mask proposals and fine mask refinements \cite{wang2023cut}, the segmentation model becomes robust to domain shifts. The detector, without any fine-tuning, is robust enough to find objects of various styles from watercolor, clip-arts to comics. The binary mask extracts multiple objects that belong to the foreground or the background using cosine similarity measures on features from models like self DIstillation with NO labels (DINO), a self-supervised transformer that learns object semantics from global representations forming from local image patches. The detector with their novel loss refines these masks and adds undetected regions from the mask proposal step.

\begin{table}[!ht]
\centering
\begin{tabular}{|llll|}
\hline
3D Representation & Explicit                                                                             & Implicit                                                                & Parametric \\ \hline
Existing Works    &
\cite{chen2022upst,carroll2010image,sahay2015geometric,kim2013wysiwyg}    
&                                              \cite{huang2022stylizednerf,tseng2022artistic,zhang2022arf,srinivasan2021nerv,chan2022efficient, chang20223d}                    
&
\cite{pang2023jointmetro,casati2019approximate, zeidler2023bodylab, jetchev2021clipmatrix}\\ \hline
Sub Types         & \begin{tabular}[c]{@{}l@{}}Point clouds\\ Voxel grids\\ Mesh structures\end{tabular} & \begin{tabular}[c]{@{}l@{}}Neural Radiance Field\\ Gaussian Splatting\\ Signed Distance Field\end{tabular} &\begin{tabular}[c]{@{}l@{}} Skinned Multi-Person\\ Linear Model       \end{tabular}\\ \hline
Visualization     
& 
\begin{minipage}{.15\textwidth}
      \includegraphics[width=\linewidth]{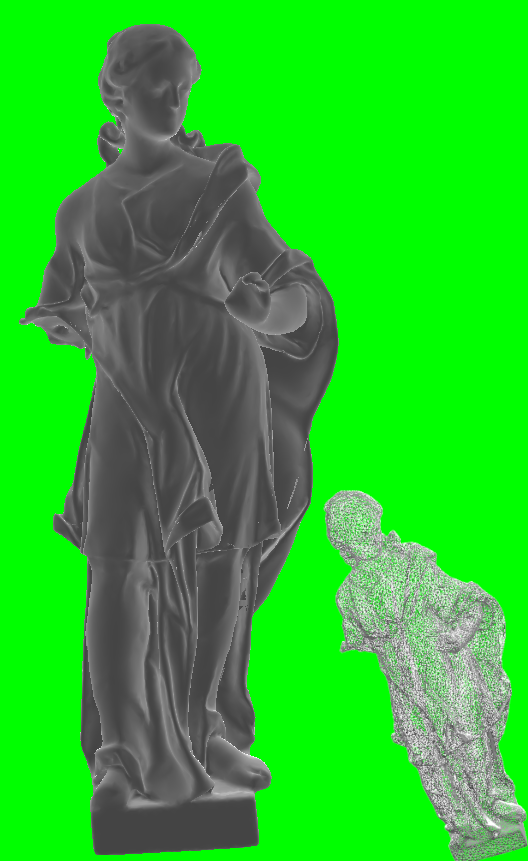}
\end{minipage}
&              
\begin{minipage}{.18\textwidth}
      \includegraphics[width=\linewidth]{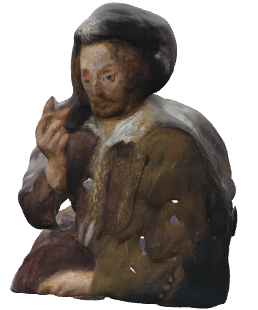}
\end{minipage}
&
\begin{minipage}{.23\textwidth}
      \includegraphics[width=\linewidth]{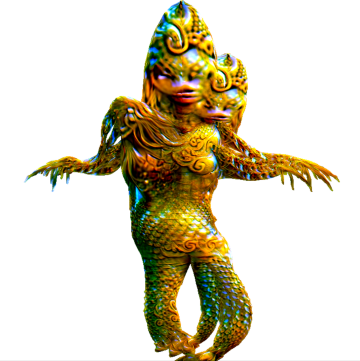}
\end{minipage}
\\ \hline
\end{tabular}
\caption{The extraction of 3D geometric data in the form of parametric representations, explicit and implicit surfaces. The table mentions the papers under these categories along with the representation types used in the subsections under 3D Features. The explicit model is a triangular mesh of sculpture and its wireframe representation. The implicit model is a Gaussian splatting rendering from a single view using the DreamGaussian architecture. Finally, the SMPL visualization is a rendering from a Contrastive Language-Image Pre-Training (CLIP) based text prompt of an Asian mermaid.}
\label{table:extraction_3d}
\end{table}

Going forward, more advanced input and output features could enhance segmentation performance. Multimodal image input features \cite{Sizyakin2020} could be extended beyond artistic object segmentation, as they derive from various data sources and feature extractions, and compensate for missing information across modalities. Additionally, they address issues like low-contrast cracks in photographs, absent cracks in IRP, and noise in X-ray images. Incorporating pseudo segmentation maps as an output feature has proven effective, leveraging discriminative features via Class Activation Maps. This approach not only improves image localization and detection simultaneously in \MV{paintings} \cite{Milani2022} but also holds potential for adaptation in human segmentation.

%MCNC uses morphological filtering from image processing to select binary crack map images that act as patches localizing cracks to condition the fully connected layer in the CNN. \HS{see if you can add more ciatations.}

\subsection{3D Features}
The feature representation of 3D models can take %the form of 
generic forms such as implicit neural structures and explicit geometric structures or specific versions from parametric constrained deformable models. Implicit neural networks represent 3D shapes and surfaces in a pointwise manner with their learned function, while Explicit models represent the objects as a 3D point collection and differ in usage by the efficiency and ease of use of their data structure for different tasks. Parametric models, on the other hand, use a small number of shape and pose parameters to efficiently represent objects in a class like Skinned Multi-Person Linear (SMPL) for human bodies with strong constraints to prevent large structural deviations. These representations are summarized with visualizations in Table~\ref{table:extraction_3d}.

\subsubsection{Implicit Models}

%nerf and style transfer module, 2d to 3d
Implicit geometry structures can separate the natural scene geometry from artistic stylization by utilizing a two-stage model with a Neural Radiance Field (NeRF) and a 2D stylization decoder which gets the projected view to style \cite{huang2022stylizednerf}. The desired style can be customized by conditioning the latent code that is the input to the decoder which also serves to deblur the rendered scene from the NeRF depth output. The model requires multiple stages \cite{tseng2022artistic} for view projections and style transfer to mix their outputs to get a stylized scene. When transferring style onto a mesh, unseen style inputs result in blurry reconstructions with naïve methods transferring only the overall color tone of the style image. By rendering the geometry and stylistic aspects separately, the ARF paper \cite{zhang2022arf} showcases the transfer of the subtle textural details of the watercolor feather image onto their Family statue scene example.

%optimization
To optimize the transfer process \cite{zhang2022arf} for better representations in view extraction, the image level style transfer uses a deferred back-propagation at a patch level to accumulate over all the patches at the neural field. To reduce computational complexity, the final model can only consider the components in the visible field of view \cite{srinivasan2021nerv} but still requires all direction illumination and material information. Alternatively, better representations can be learned using an image generative model like a Style-GAN \cite{chan2022efficient} and fed to the NeRF module, facilitating detailed feature extraction while conditioning the generator on geometric priors such as pose. For specific types of extractions such as human sculptures, implicit models like Pixel-Aligned Implicit Function (PIFu) that are trained on human data produce better results while accounting for domain shifts by adapting the intermediate features with a Maximum Mean Discrepancy (MMD) loss that aligns their moments and handles topology shifts \cite{chang20223d}.

%nerf and other 3d structures - hybrid

%misc models- sf etc
\subsubsection{Mesh Structures}

Direct 3D representative models such as voxel grids \cite{chen2022upst} ignore object artifacts found in the neural field representations in two-stage scene stylization. These are limited by the resolution of the extracted 3D model and the computational size of the intermediate features. Perspective changes warp the projected image from moving the lines of convergence constraints \cite{carroll2010image}, thereby changing the vanishing point. They achieve this view emphasis by warping the quad mesh and the corresponding homographic matrices while constraining the projection geometry.

It is possible to extract more faithful 3D models using a dictionary of surface gradients and exploiting the symmetry of such mesh structures in other views. The utilization of such self-similarity with inpainting in point clouds \cite{sahay2015geometric} finds applications in reconstructing damaged and structurally deformed architecture and sculptures.

%stereoscopic images
Stylistic renderings of multiple views for stereo paintings can use 3D paint strokes on top of partial grid mesh structures as a two-stage model \cite{kim2013wysiwyg}. 

\subsubsection{Parametric Models}
Artistic 3D models such as sculptures closely resemble the human figure, but are limited in dataset size or consist of larger variations in pose or structure to provide emphasis to foreground or background characters, sometimes exaggerating the shape from certain viewpoints. The SMPL model provides a minimal, resilient pose and shape representation through low-dimensional vectors. The customizability of intrinsic parameters accounts for the anatomical differences in artistic statues, making it handy for reconstructing statues like the Wounded Amazon with a different sized arm or for those with missing limbs such as the Esquiline Venus \cite{fu2020fakir}. By adding a Signed Distance Field (SDF) to check the occupancy of particle effect selected by the artist, the body can be textured and locally manipulated by keeping track of the normals and tangents on the SMPL's deformable mesh \cite{zeidler2023bodylab}. Instead of manual deformation and texturing, the SMPL model can be simply extended with a CLIP loss to make the mesh representation similar to that of text control with a differentiable renderer \cite{jetchev2021clipmatrix}. 
Integrating these joint interactions and their confidence coefficients with transformers increases the accuracy of human reconstruction and the speed of the mesh extraction \cite{pang2023jointmetro}. The model works well when the character images are clear, but failing in abstract \MV{paintings} in Picasso's works where parts of the head are missing or incomplete. When the body parts are occluded, it leverages joint relations to recover the skeleton topology and pose but encounters significant errors, particularly with parts like feet that have fewer adjacent joints. They use a High-Resolution Net (HRNet) to extract these human features. Representative keypoints from all of its output keypoints (e.g. nose representing the face) are then selected and the model fuses the joint and mesh information using a graph transformer model. In 3D scene extraction, template skeletons can be conditioned with bas-relief geometry, contours and silhouette information, for particular styles of sculpting to estimate 3D skeletal poses from 2D poses \cite{casati2019approximate}. The choice of the 3D mesh, such as B-mesh that provides good deformation for animation and edge flow, is separated from the rest of the scene while keeping distance information to jointly model trees, animals, and environment cues like drapery that are commonly found in these bas-relief sculptures. 

\subsection{Effectiveness of Geometry-based Methods in Extraction}

\begin{table}[]
\resizebox{\columnwidth}{!}{%
\begin{tabular}{|l|l|l|l|l|}
\hline
{ Method}                     & { Dataset}                                                                                                      & { Task}                                                             & { Metric: Value}           & { Source}                                                     \\ \hline
{ MMD and PiFU}               & { \begin{tabular}[c]{@{}l@{}}ScanTheWorld \\ scrapped meshes\end{tabular}}                                      & { \begin{tabular}[c]{@{}l@{}}3d model \\ extraction\end{tabular}}   & { Chamfer Distance: 0.047} & { \cite{Chang2022}}                          \\ \hline
{ Styled Deeplabv3}           & { \begin{tabular}[c]{@{}l@{}}Neural Style Transfer\\  on the Baidu People \\ Segmentation dataset\end{tabular}} & { Segmentation}                                                     & { IoU: 74.9\%}             & { \cite{doi:10.2352/EI.2022.34.13.CVAA-169}} \\ \hline
{ Multi-style feature fusion} & { LTLL}                                                                                                         & { Object detection}                                                 & { Accuracy: 90.9\%}        & { \cite{Ufer2021}}                           \\ \hline
{ Styled HRNet}               & { ClassArch}                                                                                                    & { \begin{tabular}[c]{@{}l@{}}Human pose \\ estimation\end{tabular}} & { mAP: 49.4\%}             & { \cite{PRATHMESH2022}}                      \\ \hline
\end{tabular}%
}
\caption{This table represents methods in the surveyed geometric feature and data extraction papers, showcasing their datasets, tasks, highest performance measure score, and sources.}
\label{tab:extraction_bench}
\end{table}

Object-centric tasks benefit from input region proposal selection strategies and additional geometric labels to add context to the input or act as pseudo-ground labels. Region selection by voting helps find and localize small motifs, achieving a maximum retrieval performance of 91.3 mAP for the LTLL dataset where other models suffer from selecting regions with low correspondences \cite{Ufer2021}. Mixing of regions selected without overlap and missing content helps in data augmentation for small datasets for logo detection for an improvement of 7.05\% mAP \cite{Fuertes2022}. Encoding the context instead of missing regions from cropping helps improve fine-tuning object detectors by 3.5\% mAP for unseen categories with an additional 2.5\% for seen categories \cite{Madhu2022}.  Leveraging this prior knowledge of pre-trained models and semantically aligning them with a newer domain helps improve performance even with difficult subdomains such as abstract paintings \cite{Lu2022}. Geometrically enhanced annotations also enhance the quality of the training dataset with maps enhancing salient regions \cite{Sizyakin2020} while eliminating irrelevant areas in crack detection or time-specific label predictions for object detection in paintings \cite{Marinescu2020}.
 
Human-centric labels such as facial landmarks and hand or body pose differ in depictions in the \MV{paintings} domain, resulting in poor results with simply fine-tuning the model on the task. Without accounting for style, body pose estimators reach less than 60\% AP \cite{Nicholas2016} while face detection algorithms reach less than 35\% F1 score on modern face datasets \cite{Wechsler2019}. When stylizing images without modifying the corresponding pose, models can get an improvement of 6\% on the mAP even without labeled data \cite{PRATHMESH2022}. Style-tuned models that pre-train on stylized content and poses gain an improvement of 36.7 mAP for the specific task of pose estimation and 34.5 mAP for the more generic person detection task. In image retrieval, performing geometric verification after fast annotation matching retrieves a longer sequence of visually similar links as compared to other models that simply match feature embeddings.
 
Parametric model-based extraction methods utilize prior knowledge to account for lower-quality data, modeling complex environments with missing information, and leveraging geometric information in the modeled latent space. They provide additional benefits in terms of a reduced computation time due to the strong prior with FAKIR \cite{fu2020fakir} extracting each iteration of a modeled statue in 9s. Additionally, it provides precise joint positions and bone radii with better shoulder location estimates compared to its counterparts, thereby producing geometrically consistent \MV{artistic 3D} models. JointMETRO \cite{pang2023jointmetro} also achieves painted sculpture reconstruction despite occluded human poses for incomplete models by utilizing this prior knowledge of human body joints. While the parametric model provides an alternative to explicit ground labels, other techniques such as domain adaptation can build upon them to reconstruct higher-quality sculptures with a Chamber Distance of 0.04 \cite{Chang2022}.

\section{Discriminative Geometric Features Analysis}
\label{section:analysis}
%data level instead?
%numerical vs ml model for the analysis n number
%understanding human response(to parts of the image eg blood or nudity to different groups), intended effect, analysis - spectrum, measuring
%topic of categorization
%measurement of trajectory(eye), metrics

%add in image retrieval - relationships

% \begin{table}[ht]
% \centering
% \begin{tabular}{|l|lll|l|l|}
% \hline
% \multirow{2}{*}{Model level} & \multicolumn{3}{l|}{Geometric features}                                    & \multirow{2}{*}{Transformations} & \multirow{2}{*}{Effect} \\ 

%                              & \multicolumn{1}{l|}{Implicit} & \multicolumn{1}{l|}{Explicit} & Relational &                                  &                         \\ \hline
%                              & \multicolumn{1}{l|}{}         & \multicolumn{1}{l|}{}         &            &                                  &                         \\ \hline
% \end{tabular}
% \label{geom_analysis_overview}
% \caption{An overview of analysis using geometry information at different levels and settings and their effects.}
% \end{table}
%rewrite
%Various painting classification tasks can utilize geometric information in the form of classifier extracted geometric features or relationships from spatial correspondences encoded as graphs. 
Various painting analysis tasks utilize geometric features from low level (\emph{i.e., local feature descriptors such as brushstrokes or optical flow maps encoding direction} )  and intermediate level (\emph{i.e., cross spatial correspondences between objects to identify keypoints and landmarks}). These analysis methods even utilize outputs from feature extraction methods such as a list of bounding boxes. These geometric features and data are utilized in the tasks that are sub-categorized into scene classification, retrieval and style classification.

Scene classification relies on similarity measures to determine object arrangement and assign scene labels. It involves three core stages: feature extraction 
 (e.g ResNet without fully connected layers), spatial correspondence encoding 
 (e.g Attention, K-means clustering), and output mapping (e.g. Aggregation of bounding boxes, multiclass classifiers with Softmax activations). In contrast, style classification focuses on identifying unique styles in \MV{artistic images}, which can vary in visual cues rather than content. Notably, stylistic manipulations in images can impact feature extraction and geometry due to object detectors, posing challenges on both scene classification and retrieval. Scene retrieval aims to find images resembling a reference by mapping output features to identify closely matching candidate images of a specific scene class.

%find scene classification survey paper

%not yet task, image comparison - this is a data structure/feature/latent feature/img representation effectively
%citations(no dl) - liu2021 here
\subsection{Object Detection}
Objects in paintings have large shape exaggerations in modalities such as cartoons, vary drastically in their composition with cluttered scenes consisting of objects of different scales and spatial arrangements. The spatial layouts are crucial in scene understanding \cite{Madhu2022, Milani2022}, with objects representing visual motifs for artists or indicating time periods and culture by their co-occurrence with other objects. These artistic datasets are small in size \cite{ahmad2023toward}, with some providing only image-level annotations \cite{Thomas2018, seo2016video} or missing object-level labels \cite{Madhu2022, Milani2022}.

Traditionally, object detectors in landscape scenes \MV{focused on analyzing and understanding low-level features, such as brushstrokes, for capturing scene dynamics using optical flow although the result can be suboptimal when noisy object regions are not effectively detected as principal components. By integrating these low-level details with region-based segmentation algorithms like Comaniciu’s mean shift clustering, which groups input scenes by color, the system provided a deeper understanding of the composition at the object-level. The segmentation information encoded by the clustering method contains the region information distribution according to the object thereby capturing the variations in appearances and their relationship with each other through. This data-driven approach allowed for more refined scene interpretation when fed into a threshold-based classifier, facilitating a clearer distinction between objects and background elements  \cite{seo2016video}.} More recent computer vision techniques used models pre-trained on a larger domain for the same classes in a target domain by fusing the style from the \MV{artistic target image} modality and the content of the source modality. To create the synthetic pair, methods like arbitrary style transfer using Ada-IN, which learns a style transformation network to translate images from one domain to another \cite{Thomas2018}. Such a method provided easier access from faster training and no fine-tuning to multiple sub-tasks (when considering multiple modalities in domain adaptation) is needed, unlike learning generative models such as GANs. The work shares the pre-trained backbone and fully connected classifiers with multinomial logistic losses from a domain confusion loss to predict the domain of the image and an object classification loss. Multiple modalities force the network to learn a general representation, enforcing style invariance with the choice of style transfer affecting the retained structure and details of the synthetic image. Realistic paintings get limited improvement while modalities that emphasize shapes with their contours like cartoons and sketches gain substantial performance gains.

When the feature alignment process between domains uses a generative model such as Cycle-GAN \cite{pasqualino2022multi} instead of style transfer, it loosens the requirement of pairwise source and target domains for image translation. It also provides a fully differentiable domain adaption \MV{method where the multiscale detector, Retina-Net, acts as another discriminator for multiple adversarial losses, one for domain confusion and the other for object prediction to understand the variations in artistic content across domains while constraining the learned transformation to produce an object of the target style. In \MV{galleries}, these models are utilized to study how variations in lighting conditions, viewpoints, and mixed environments affect the model’s ability to correctly interpret and adapt to painting or sculpture regions. The interaction between \MV{artistic images or 3D models} and real-world surroundings creates complex and varied input data, and analyzing these variations helps reveal strengths and weaknesses in the model's alignment process. While the model excels at translating artistic styles and maintaining target domain features, it struggles with scenes that involve multiple \MV{artistic images ranging from paintings, clip-art and comics}, where the co-occurrence of certain object classes can mislead the interpretation of the discriminative regions. Effective analysis of detected objects depends on the quality of feature alignment from the source and target domains, particularly when the dataset comes from varied \MV{artistic image} modalities \cite{ahmad2023toward}.} By modeling parallel object proposal networks, the classifier can better handle variations in data from fusing regions and adjusting its parameters based on the distribution through XGBoost. The boosting algorithm helps emphasize more difficult \MV{or rare cases, which is crucial for understanding less frequent objects or features within the dataset. However, despite the model’s strong performance in multi-scale analysis and fast inference provided by YoloV5, it often struggles with datasets heavily focused on people, where searching for other, (e.g. ess common) objects becomes more challenging} without further modifications to the pipeline.

\subsection{Style Classification}
Style classification involves artist identification and the common visual elements, techniques and forms used in their works. The artists attribute the forms to lower-level textures, such as their choice of color palette, brushstroke, or materials, up to the higher-level choice of \MV{fine-art painting compositions}. Style classifiers benefit from feature fusion techniques that merge geometric image representations with deep learning features as input to a multiclass classifier. These geometric representations are handcrafted for the problem to account for the large inter-class variation in styles and class imbalances stemming from artist-based classification, which result from variations in the artists' prolificity. In older works, CNNs, \MV{were used purely as object feature extractors which is less effective in capturing image representations compared to learned ensembles of handcrafted features like Classemes or PiCoDes \cite{saleh2015large}. However, their performance improved significantly when the object region was first extracted and used as input to the model. This approach highlights the importance of isolating relevant regions for analysis, enabling CNNs to better understand and represent the essential characteristics of the image, thus offering a more accurate interpretation.} When a DPM provides class-specific regions to a multiscale CNN to provide a holistic encoding and learn a distribution of local encodings through a GMM, their joint embedding after aggregation through techniques like Fisher vector gives better performance \cite{anwer2016combining}. More recent work has enhanced CNNs' ability to analyze and understand image representations by incorporating discriminative signals from an SVM-based classifier. \MV{This approach refines the clustering criteria, allowing the model to generate centroids that more closely align with the original target label distribution \cite{sandoval2021adversarial}. This combination of deep learning and SVM-based analysis facilitates a deeper understanding of the underlying data structure, ensuring that the representations captured by CNNs better reflect the true characteristics of the target labels.}

\subsection{Scene Classification}
Scene classification involves the general subject matter or the semantics in the painting and considers categories like outdoor and indoor-based scenes, landscapes and portraits, seascapes and landscapes, still life or other labels describing the scene type. Simple methods like segmentation (eg. Normal Cuts) can extract visual descriptors like HOG or GIST \MV{from regions} at the image level \cite{condorovici2013painting} \MV{for representing the structure and texture within each segmented region.} These object descriptors are used as %form 
the input to a Bayes classifier to form the RoI pooling operation in a multistage object detection model. Due to the simplicity, the classifier focuses on colors and results in mistaking images from nudes and portraits since they both contain skin colors or they can not distinguish between cityscapes and landscapes due to the latter being the superset. Later models use CNNs to represent generalizable feature representations for multiple modalities alongside constraints like MMD that force a shared representation among different CNN heads \cite{castrejon2016learning}. With modality-specific fine-tuning, the target dataset can have a smaller number of samples while the distribution constraint enables an emergent alignment of objects shared across multiple representations. 

\subsection{Human Perception Analysis}
\label{section:perception}
Objects in paintings can appear distorted despite being portrayed with the correct geometry based on the viewer's vantage point from large visual angles that tilt and straighten, reduced saliency of peripheral objects, to depth-wise elongations \cite{Dejan2009}. Experiments involving participants to move towards the painting until they saw the object of interest take the desired shape or subtend an angle showcased similar results to that of projective geometric analysis, but to varying extents. When measuring their response, the farther vantage points had varying perceptions of distance in peripheral objects for large paintings or those approximating 3D scenes. A case study of Piranesi's painting composition hints at possible approaches to balance the trade-off between accurate scene geometry against perceptive distortion \cite{rapp2008geometrical} with the pieces utilizing projections from multiple viewpoints along the central vantage line. While this reconstruction shows inconsistency using geometric restitution as a tool for analysis, it does provide a view to different proportions, sizes and relative distances of non-distorted objects from different viewing angles.

\subsection{Effectiveness of Geometry-based Methods in Analysis}
\begin{table}[]
\resizebox{\columnwidth}{!}{%
\begin{tabular}{|l|l|l|l|l|}
\hline
 
{ Method}                                                                                  & { Dataset}                  & { Task}                       & { Metric: Value}                  & { Source}                                        \\ \hline
 
{ \begin{tabular}[c]{@{}l@{}}Faster R-CNN with CAM \\ and ResNet-50 backbone\end{tabular}} & { ArtDL 2.0}                & { Object Detection}           & { mAP: 41.5\%} & { \cite{Milani2022}}            \\ \hline
 
{ MLCNN}                                                           & {\begin{tabular}[c]{@{}l@{}} Artsy,\\ WikiArt paintings\end{tabular}} & { Orientation classification} & { Accuracy: 92.42\%}              & { \cite{zhao2023research}}      \\ \hline
 
{ DPM detector}                                                    & { painting-91}              & { Style classification}       & { Accuracy: 74.8\%}               & { \cite{anwer2016combining}}    \\ \hline
 
{ GMM}                                                             & { CMPlaces}                 & { Scene retrieval}            & { mAP:14.2\%}  & { \cite{castrejon2016learning}} \\ \hline
\end{tabular}%
}
\caption{This table represents methods in the surveyed discriminative analysis papers, showcasing their datasets, tasks, highest performance measure score, and sources.}

\label{tab:analysis_comparison}
\end{table}

In the task of cross-domain object detection, additional representations such as CAM \cite{Milani2022} or context encoding \cite{Madhu2022} can compensate for a lack of ground truth (GT) label while bringing in contextual information as learned by pre-trained models on a larger, well-annotated dataset. CAM acts as a pseudo-GT beating the SOTA on the weakly supervised object detection task in the IconArt dataset by 14\%, while with the context encoding the finetuned model beat the SOTA in the mean average precision (mAP) by about 3.5\% at 0.25 intersection over union (IoU) for UnSeen categories.
 
For the broader task of scene and style classification, the content of the painting has the greatest effect on the classification accuracy favoring methods that process the local and global regions separately before integrating the results. For example, the number of classes from both painting categories and the number of directions can be reduced to an umbrella, holistic classes to increase the average accuracy to 90\% for orientation classification \cite{zhao2023research}. In the latter task, part-based models or models that extract text and style separately improve model accuracies by 6.4\% and 13\% respectively, compared to the non-contextual cases. \MV{This approach allows a more detailed understanding of the distinct elements within the data, enabling the model to better capture and interpret the contextual relationships between features, and thus leading to more accurate predictions.} In other cases, the feature representation can be conditioned to include context such as meta-data \cite{fumanal2023artxai} for an improvement of 26\% as compared to a 6\% increase from building upon context-aware solutions.

\section{Synthesis with Geometric Features}
\label{section:synthesis}
% \HS{This section should be about generating things (e.g. images, parts of an image, novel views, etc.) with the use of any geometric features.}

\begin{sidewaystable}[]
\centering
\begin{tabular}{|lllll|}
\hline
Synthesis Task      & Style Transfer                                                                                                                                 & Inpainting                                                                   & Relighting                                                                 & Conditional Image Generation                                                    \\ \hline
Existing Papers     & 
\begin{tabular}[c]{@{}l@{}}
 \cite{liu2022geometric, yang2022industrial, liu2021learning, vulimiri2021integrating, nakano2019neural, geng2022ptgcf, papari2009glass, yin2019instance, kim2020deformable, kopanas2021point, Qingfu2020, liu2020geometric, vulimiri2021integrating, alexandru2022image, Du2022, upadhyay20223dstnet}
 \end{tabular}
 & 

 \begin{tabular}[c]{@{}l@{}}
\cite{sahay2015geometric, bird2021continuation, ciortan2021colour, Cipolina-Kun_2022, zhang2023adding}
\end{tabular}
&
\cite{Stork2006, chen2012artistic, chen2021perspective, chen2012artistic, Henz2017, Henz2014ImageRU, Mishra2022CLIPbasedNN, Jin2022LanguageguidedSS, zhao2020painting, tan2015decomposing, koyama2018decomposing}
    & 
    \cite{hou2021mw, hou2021mw, huang2023controllable, abrahamsen2023inventing, chang2023design, chen2023conditional, peng2023difffacesketch, yang2022industrial, zeng2023scenecomposer}

    \\ \hline
Types of techniques & \begin{tabular}[c]{@{}l@{}}Geometric priors\\ Model induction via \\ perceptual losses\\ Post-processing and \\ style refinement\end{tabular} & \begin{tabular}[c]{@{}l@{}}Mask-based editing \\ Structure completion\end{tabular} & \begin{tabular}[c]{@{}l@{}}Color transfer\\ Portrait lighting\end{tabular} & \begin{tabular}[c]{@{}l@{}}Generative Adversarial \\ Network\\ Variational Autoencoder\\ Diffusion model\end{tabular} \\ \hline
Inputs 
&
\begin{minipage}{.1\textwidth}
      \includegraphics[width=0.9\linewidth, height = 1.1\linewidth]{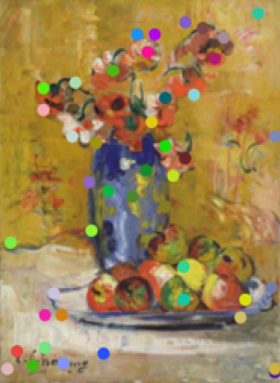}
      source\\
      \includegraphics[width=0.9\linewidth]{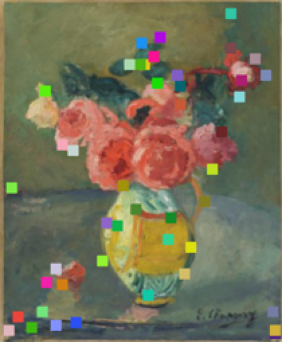}
      reference
\end{minipage}  
&
\begin{minipage}{.1\textwidth}
      \includegraphics[width=1.3\linewidth , height= \linewidth]{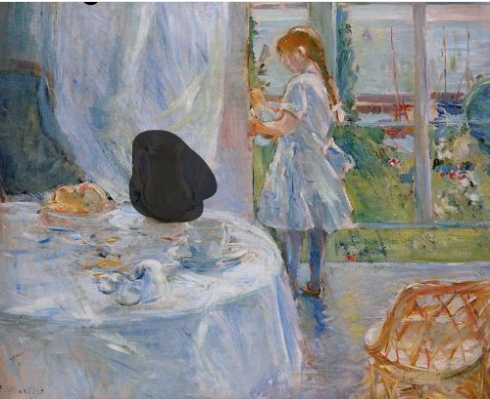}
      source\\
      \includegraphics[width=0.9\linewidth , height= 1.1\linewidth]{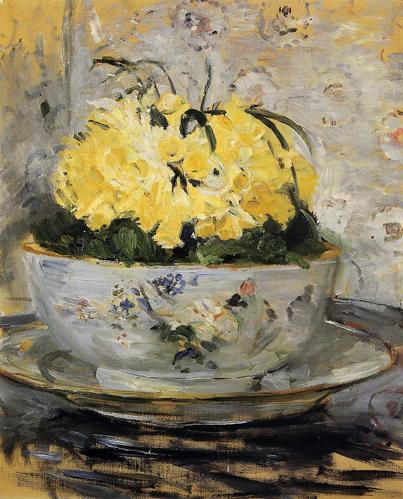}
      reference
\end{minipage}  
&
\begin{minipage}{.1\textwidth}
      \includegraphics[width=\linewidth]{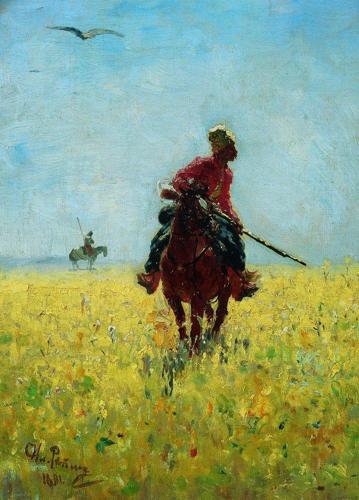}
      source
\end{minipage} 
&
\begin{minipage}{.07\textwidth}
      \includegraphics[width=1.3\linewidth , height= 1.7\linewidth]{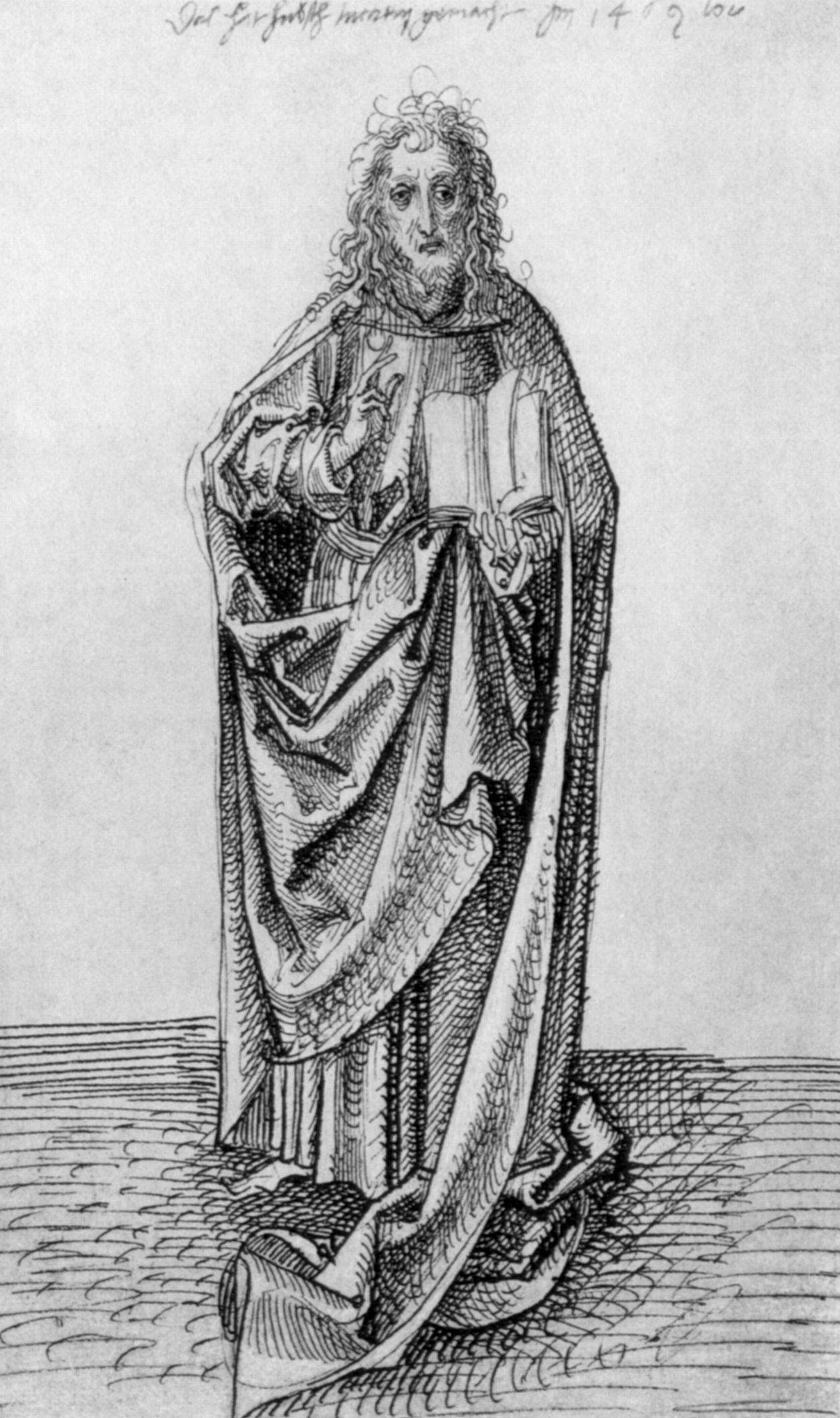}
      source
      \\
      \includegraphics[width=1.3\linewidth , height= 1.5\linewidth]{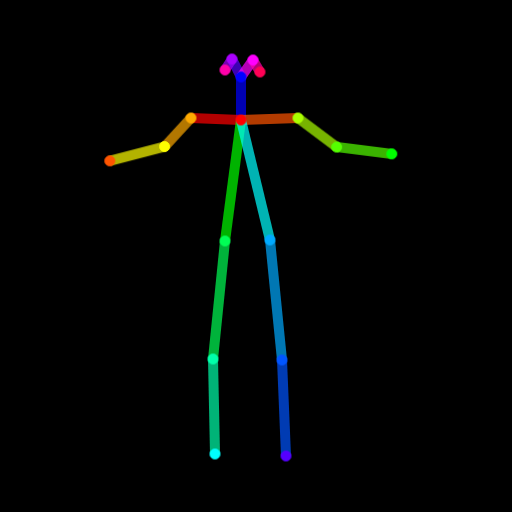}
      pose
\end{minipage}                                                                     
\\ \hline
Outputs             
&

\begin{minipage}{.1\textwidth}
      \includegraphics[width=0.9\linewidth]{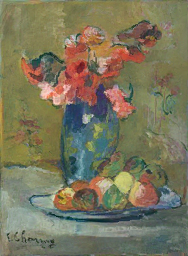}
\end{minipage} 
&
\begin{minipage}{.1\textwidth}
      \includegraphics[width=1.3\linewidth, height= 1.2\linewidth]{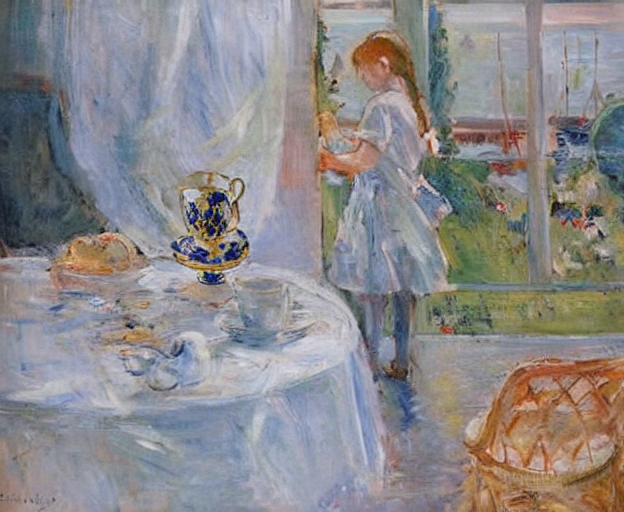}
\end{minipage}  
&
\begin{minipage}{.1\textwidth}
      \includegraphics[width=\linewidth, height= 1.2\linewidth]{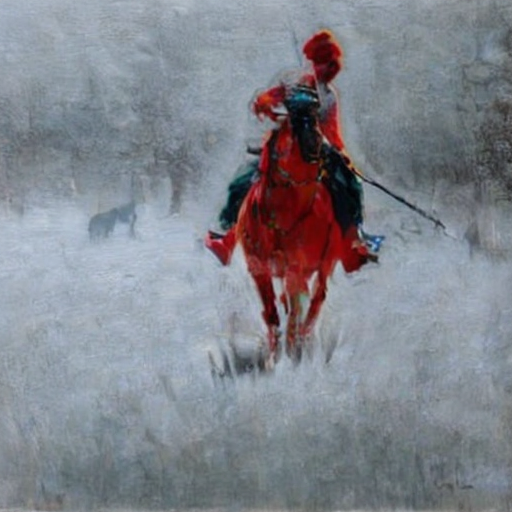}
\end{minipage} 
&
\begin{minipage}{.1\textwidth}
      \includegraphics[width=0.9\linewidth , height= 1.2\linewidth]{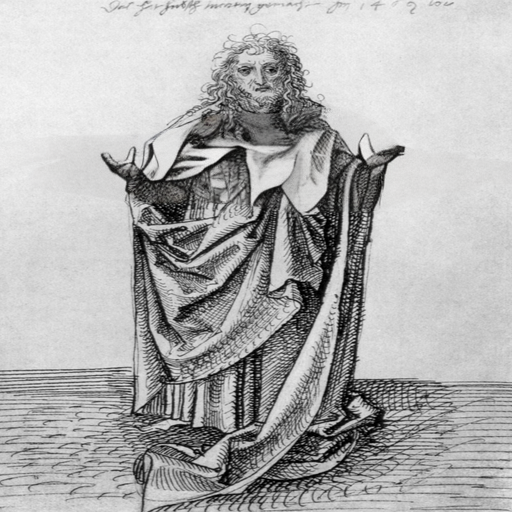}
\end{minipage}  
\\
\hline

\end{tabular}

\caption{The table lists image synthesis methods for image stylization, manipulation, relighting and generation with their inputs and outputs. It also lists the techniques used for these tasks with their papers in the Synthesis section. Style transfer involves the deformation of the flower vase still life images in reference to that of the source. Inpainting on the other hand fills in the missing region indicated by the mask in the impressionist-style source image with that of the reference still life image. It results in the source image of a lady near a dining table replacing the mask with a similar flower vase from the reference image, but one matching the blue color tones of the source. Relighting changes the foreground and background lighting of a source image from a cavalry riding through a meadow in spring to winter. The Image generation task is visualized here as conditionally regenerating a sketch in the pose of the reference image. It changes the man preaching with a book to one with spread arms.}
\label{table:synth_main}
\end{sidewaystable}

The synthesis section covers the generation and manipulation of \MV{artistic images or 3D models} ranging from paintings, cartoons, sketches to sculptures. The generative models include components to separate style from structure, including geometric deformation modules or networks along with a separate module to blend the separated components together. Part of the model store or use the geometric feature as input to a different part of the generative pipeline. Geometric data in the form of masks and semantic maps, 3D representations such as point clouds or polygon meshes, are used as additional input to these models. In image manipulation, some regions of the image are changed by adding and removing objects or to match the deformations as in a reference image. The task of novel view synthesis covers unseen \MV{artistic 2D/3D data} sampling using image generation models, relighting to provide a view of the image with different lighting, time lapses or seasons, and rendering to change the local geometric details. These main techniques are summarized with visualizations in Table \ref{table:synth_main}.

The synthesis section covers the generation and manipulation of images or 3D models of \MV{artistic data} ranging from paintings, cartoons, sketches to sculptures. 

%2 more papers to be added
\subsection{Image Manipulation}
Image manipulation involves deforming the contents of the image to that of a reference image or editing the objects into or out of the source image. The objects immersed in the scene can belong to the same or different \MV{artistic} modality while the region deleting the object uses the neighbouring area to fill in candidate outputs. The main challenges during the manipulation of artistic images are artifacts around the boundary or bad correspondence matches indicating a semantic gap.

%content preserved
\subsubsection{Style Transfer }

\begin{sidewaystable}[]
\centering
\begin{tabular}{|ccllll|}
\hline
\multirow{2}{*}{Image Domain} & \multirow{2}{*}{Content Image} & \multirow{2}{*}{Style Image} & \multicolumn{3}{c|}{Stylized Image}              \\ 
 &
   &
   &
  \multicolumn{1}{c}{\begin{tabular}[c]{@{}c@{}}Neural Style \\ Transfer\end{tabular}} &
  \multicolumn{1}{c}{\begin{tabular}[c]{@{}c@{}}Adaptive Instance\\  Normalization\end{tabular}} &
  \multicolumn{1}{c|}{\begin{tabular}[c]{@{}c@{}}Deformable Style\\  Transfer\end{tabular}} \\ \hline
Same Domain   & 
\begin{minipage}{.15\textwidth}
      \includegraphics[width=\linewidth]{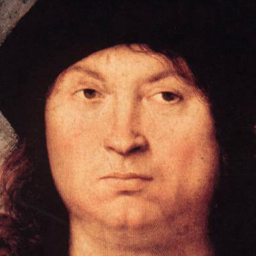}
    \end{minipage}
    & 
    \begin{minipage}{.15\textwidth}
      \includegraphics[width=\linewidth]{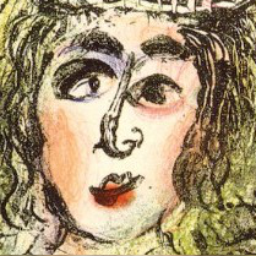}
    \end{minipage}
    & \multicolumn{1}{l}{
    \begin{minipage}{.15\textwidth}
      \includegraphics[width=\linewidth]{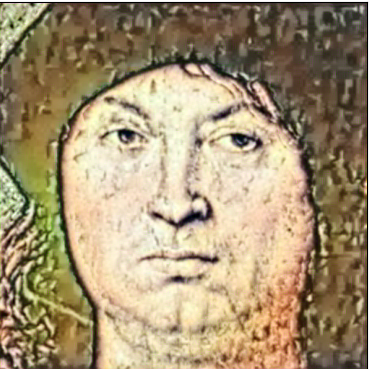}
    \end{minipage} }
    
    & \multicolumn{1}{l}{
    \begin{minipage}{.15\textwidth}
      \includegraphics[width=\linewidth]{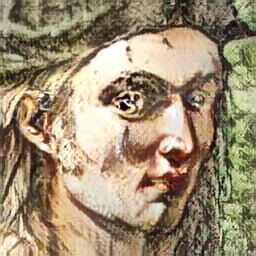}
    \end{minipage} 
    } &  
    \begin{minipage}{.15\textwidth}
      \includegraphics[width=\linewidth]{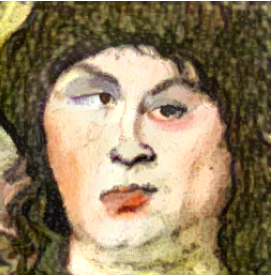}
    \end{minipage}\\ \hline
Cross-Domain &  \begin{minipage}{.15\textwidth}
      \includegraphics[width=\linewidth, height =1.1\linewidth]{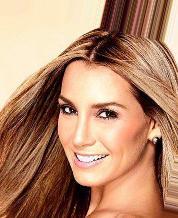}
    \end{minipage}
    & 
    \begin{minipage}{.15\textwidth}
    
      \includegraphics[width=\linewidth, height =1.1\linewidth]{original_figures/style_transfer_examples/84.png}
    \end{minipage}
    & \multicolumn{1}{l}{
    \begin{minipage}{.15\textwidth}
      \includegraphics[width=\linewidth, height =1.1\linewidth]{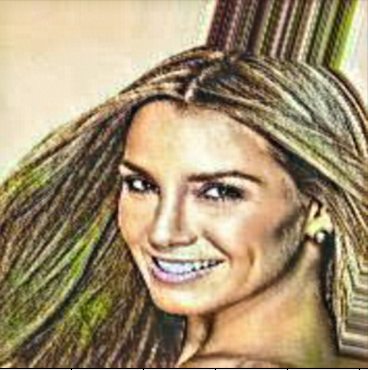}
    \end{minipage} }
    
    & \multicolumn{1}{l}{
    
    \begin{minipage}{.15\textwidth}
    
      \includegraphics[width=\linewidth, height =1.1\linewidth]{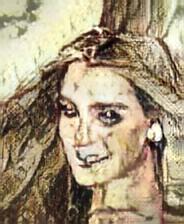}
    \end{minipage} 
    } &  
    \begin{minipage}{.15\textwidth}
      \includegraphics[width=\linewidth, height =1.1\linewidth]{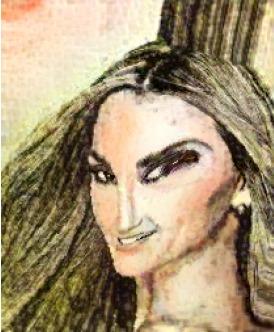}
    \end{minipage}\\ \hline
\end{tabular}
\caption{Visualization of stylized outputs from different style transfer techniques. The style and content images are either from the same \MV{2D/3D artistic data} domain or the content is from real-world images to represent the case of using style transfer as purely a data augmentation technique. These style transfer techniques are covered in the Synthesis section and show the difference in outputs from the simple iterative style transfer method to one that uses the style transformation network and finally, one that is geometrically aware. The style of the abstract depiction of the lady's face is muted for the first, while the correspondence between the facial features is preserved in the 2nd style transfer output. When the geometric deformations of the style are also taken into consideration by matching correspondences, the source images of a painting or photo of a person match the abstractness of the style.  }
\label{st_vis}
\end{sidewaystable}

%texture
%independence of style and geometry for control in editing
Style transfer offers the benefit of considering the holistic geometry as compared to traditional image processing techniques at the cost of smooth and artifact-free stylizations. \MV{Artistic 2D and 3D data} can benefit from this separation since it often geometrically deforms images as a stylistic choice. It can be formulated as optimization or transformation of the stylization and geometric modules at the pre-processing, model-level and post-processing stage.

%style transfer preprocessing
%geometric optimization
During style transfer pre-processing, we learn the distribution of shapes \cite{liu2022geometric} while promoting textural invariance using augmentation to improve object detector performance by tackling implicit model bias. However, there is overlap through style per-training to projection or shapes and geometric augmentations from deformations or distortions of size and orientation. We can pre-process style transfer in the extracted mask for the structural warp \cite{yang2022industrial}, but we need to refine the mask boundaries for better texture transfers. We could also embed class-specific warp feature fields through style content image pairs and their corresponding vector field \cite{liu2021learning}. These warp fields account for intra-domain and cross-domain generalization. Alternatively, we can trade off the mesh style to structure integrity by weighting the topological optimization against the style loss\cite{vulimiri2021integrating}. These geometric deviations include unconnected objects or unfaithfulness to the overall main design, requiring post-processing filters for imposing global geometric features. 

%style transfer with geometric processing - taking halves
%SBR
The methods that use a specialized model for handling geometry in tandem with style transfer require preconditioning or pretraining and cannot handle multiple representations. Stroke-based rendering can model strokes and the image content separately with a neural painter as an image generative model and style transfer's content loss respectively\cite{nakano2019neural}. These strokes can be approximated as a joint fusion problem in the Fourier domain\cite{geng2022ptgcf} by shifting the brush stroke from the source image patch to that of the mean patch. Vector fields along with noise can also model brush stroke stylizations of the input image \cite{papari2009glass} while preserving object-level details with their contours untouched. In facial transfer, geometric flow models \cite{yin2019instance, kim2020deformable} with facial landmarks blend and warp details together with an adversarial or deformation loss model the addition and subtraction of the attribute while preserving the person's identity. 

%style transfer as post processing 
By using style transfer as a post-processing task, it can be treated as a separate optimization task on the output of an existing pipeline \cite{kopanas2021point, Qingfu2020}. This extends its application to input and output images in the 2D and 3D domains.

%2d
To preserve geometric features in 2D images, we need three images to provide the style and geometry information of the stylization, as well as the target geometry or matching algorithms to provide the correspondences at the image or model-level. Geometric Style Transfer\cite{liu2020geometric} uses a geometric warping network with a specialized style image alongside loss pairs for structure and style. When transferring style to objects detected in bounding boxes, geometry-aware style transfer can fail if there are no good semantic matches between the style and content images \cite{alexandru2022image}. Neural style transfer can synthesize designs with a mesh reference and a novel topology optimization inspired compliance loss \cite{vulimiri2021integrating} to separate the geometric information from texture. Bounding boxes facilitate the warping of specific matches of regions of the content and style images \cite{alexandru2022image} before applying style transfer for the desired texture while retaining multiple object structures. Alternatively, the selected regions can constrain the style loss, providing spatial and aesthetic effects, error correction and controlled transference of styles according to the masked regions \cite{kolkin2019style}. Thus, the structural loss can be modified with warping, reconstruction or geometric consistency terms to achieve geometric deformations and projections that match the style images.

%2d to 3d
When translating an image to a 3D model, stylized \MV{outputs} could preserve their pose after style transfer \cite{Qingfu2020} using a novel style reconstruction loss on the 2D stylized image which is lifted to the 3D depth image as a pose regression task. Arbitrary style transfer allows the stylistic transfer of the pose only, where the style loss adds a new HSV colorspace loss for color insensitive style transfer along with a slightly modified style and content loss to follow the general direction of the style and content features than a reconstruction loss. The added self-supervision loss corrects the lifted stylized pose in the 3D space when reconstructing the stylized image, thereby using the geometric nature of the bone map with style transfer. 

%3d
3D stylistic modeling uses a separate stylistic module or loss to transfer the style while preserving the geometric features in an implicit or explicit representation. By sketching the contours at an angle of a sparse point cloud \cite{Du2022}, the artist can interactively reimagine the generated 3D model from the initial models of pre-existing categories, with contours retrieved from viewpoint matrices. With the help of the pre-existing dataset, the user input contours can be matched and retrieved for that part of the viewed model. Similarly, we can represent the structure of 3D point cloud vases as features extracted \cite{upadhyay20223dstnet} to transfer shapes with style transfer losses and a laplacian loss to preserve local details such as edges, contours and patterns better.

%content not preserved
\subsubsection{Inpainting}

Inpainting in \MV{artistic 2D/3D data structures} is used in applications for image \cite{bird2021continuation} and region level\cite{sahay2015geometric} completion, editing\cite{zhang2023adding} and restoration\cite{ciortan2021colour}. They do not preserve part of the content in the reference image but transform the style of the target object to that of the source. They perform image completion in selected patches or masks.

%patches, completion
Structure completion in Inpainting incorporates the target boundary-based patch selection as a search and voting optimization task \cite{datta2017image}. This patch matching incorporates geometric transformations such as reflections, non-uniform scaling and other perspectives through affine transform approximation. Alternatively, region filling is possible from data fusion from overlapping local patches with gradient-based self-similarity \cite{sahay2015geometric}.
Using cropped patches, we can achieve image completion, which produces ringing artifacts \cite{bird2021continuation} with vanilla image generators like Generative Adversarial Networks (GANs) that follow a loss of high-frequency information. These methods help remove boundary artifacts within the patches but not on the patch edges themselves.

%masks
%guidefill 3d model - bring to mesh structures instead?
Mask-based inpainting commonly incorporates explicit geometric features such as object masks from scene segmentations \cite{doi:10.2352/EI.2022.34.13.CVAA-169}. Their flaws are mainly from the properties not included in the mask types themselves such as scene segmentations with depth discontinuities and changing boundaries from ghost pixels \cite{he2011image} that are mistakenly attributed to different objects. Inpainting specific aspects such as high-frequency details of a painting is possible through edge maps\cite{ciortan2021colour} in a different domain or colorspace. These masked images allow editing of subjects in the erased area with generative models like latent diffusion models and Large Language Model based prompt guidance~\cite{Cipolina-Kun_2022, zhang2023adding}.

%harmonization
% \cite{peng2019element}
% \cite{luan2018deep}
% \cite{bao2022deep}

\subsubsection{Conditional Image Generation}
Generative Adversarial Networks help in the quick sampling of outputs with multiple controllable attributes embedded as a conditional vector, an additional input that commonly represents a style or shape vector.  Its attributes are formulated as disentangled representations to form independent control factors, with multiple modalities (such as other input domains for \MV{artistic} modalities or geometric labels) encoded with Cycle-GAN variants~\cite{hou2021mw, huang2023controllable} to facilitate unpaired domain-wise translations and learning their correspondences. Variants of Generative Adversarial Networks can be used for synthesizing warped stylization images, for example, using facial landmarks or edges from line-art \cite{ci2018user}  as the conditioning vector in conditional Cycle-GAN. Conditioning inputs such as segmentation maps provide size and location cues, however, the model produces worse results if the semantic classes and their object appearances diverge significantly from the pretraining data distribution. On the other hand, keypoints indicate local relationships and correspondences, which help generate results with significant shape exaggeration. However, the mismatches or deformations can generate implausible results that do not match the input distribution. The choice of conditional for GAN style guidance influences the geometric embedding network's ability to capture finer, localized details like strokes through directional fields, thereby shaping the diversity of synthesized styles. \cite{abrahamsen2023inventing}. By borrowing the stylistic deformations from the source image and the reference's colors and content, the model generates samples that belong to a new synthesized \MV{AI-graphics} movement with its real-world natural input distribution. In place of external conditional inputs, the correspondences between parts of the paintings can be approximated by cycle consistency losses while additional losses such as brush stroke and ink wash losses help the model simulate geometric textures such as brush strokes or the washed-out effects of paints such as Chinese inks \cite{he2018chipgan}.

Diffusion models \cite{chang2023design} extend the level of control to different shapes and styles while also learning attributes that adapt the conditional's data distribution. These models learn an iterative mapping from a simple distribution, such as the Standard Normal, to a complex distribution. In the case of latent diffusion models, they learn the latent space's distribution to better represent and mix the input distributions. The encoder of the geometric labels \cite{yang2022industrial} such as sketches and segmentation maps are frozen and the outputs are fused with the input image embeddings. For the former, the temporal order of sketches can be encoded into another latent space with part-wise Autoencoders or Variational Autoencoders to reduce computation \cite{chen2023conditional, peng2023difffacesketch}. Compared to GAN-based methods, they can generate better results with people with accessories and facial features, as well as supporting different levels of part abstractions. Conditioning on segmentation maps \cite{zeng2023scenecomposer} provides coarse geometry that can be enhanced with other labels such as text that make them adaptable instance labelled maps. While these pseudo-labels can be weighted to control their influence in the result, the properties of the geometric label are not shared with the text descriptions without clearly defined instance boundaries. Text labels serve to offer global and semantic context clues within spatial layouts, yet they do not directly enable manipulation of the layout itself \cite{chen2024training}. This helps to clarify and rectify ambiguous and erroneous learned correlations that may arise from coarser maps. Since the encoders do not share information, the embeddings cannot learn the shape separations in the mask encoder which extends to other embeddings of geometric cues.  

\subsection{Novel View Synthesis}
Novel view synthesis in \MV{artistic image and 3D model} datasets refers to techniques that change the perspective of the scene and its objects by relighting and recolorization. This changes the focus of the subject in the scene and aligns the illumination process to be closer to real-world settings to better convey the artist's intentions. In these \MV{artistic images}, the illumination sources can drastically vary from realistic sources in terms of color, direction, and intensity which encourages modulating the lighting on the extracted geometric shape of the painting.

\subsubsection{Relighting}
%too niche? cant find papers in my collection that focus on lighting on surfaces in painting. maybe this is more perspective?
\MV{The manipulation of directional lighting in paintings \cite{Stork2006} allows art historians to gain new interpretations of the \MV{artistic images} by determining the nature and effectiveness of optical instruments in the past. Adjusting the lighting to account for cast shadows, specular reflections, and self-shadows, while incorporating point source illumination information, can reveal previously unnoticed elements of the composition. By examining the direction of illumination, new insights can emerge, such as identifying geometric inconsistencies or uncovering occluding contours, which may hint at image tampering or offer alternative realist painting compositions.} 

Simple techniques like illumination template matching \cite{chen2012artistic} help in lighting transfer from the source to the target image by warping the matched face descriptor if the light source is simplistic and there is only one subject. In interior lighting, the light sources from the style image can be transferred to the content by extracting the perspective information with key-point detection to warp the surface map according to the style elements \cite{chen2021perspective} before restoring the perspective. Similarly, face illumination descriptors on an active shape model along with deformation transformations can perform facial illumination transfer \cite{chen2012artistic}. Utilizing 3D models as shading proxies allows for lighting transfer from the user-provided light source direction \cite{Henz2017} or segmented reference object \cite{Henz2014ImageRU} to the target shading proxy. This proxy contains artistic style, brush strokes and color information to learn an implicit normal and depth map , which overcomes the limitations of previous methods on handling highly stylistic scenes. Alternatively, the use of CLIP \cite{Mishra2022CLIPbasedNN} for obtaining physical lighting properties and local geometry information can transfer lighting using explicit normals and materials. Similarly, a 3D mesh structure can be stylized according to the text prompt with CLIP guidance \cite{Jin2022LanguageguidedSS} while keeping the differentiable rendering of the correct viewpoint and lighting of the final 3D mesh.

\subsubsection{Recolorization for Artistic Time Lapse}
Artistic Time Lapse decomposes the painting into its constituent objects with the help of frames of the video creating the \MV{artistic animation} to relight them according to the environment's illumination source for a day or across seasons. The methods detect keyframes with color shifts to get the art's decomposition into layers indicating depth from the artist's perspective and predict the next frame by learning the sequencing using a Conditional Variational Autoencoder \cite{zhao2020painting} that conditions on the previous frame. To generate time lapses of the painting itself instead of the painting process, the albedo map is estimated for each frame, clustered with their linear layering to pick the colors not affected by lighting and hue shifted with the artist's choice of colors to produce the time-lapse effect \cite{tan2015decomposing}. The decomposition process can be differentiable to extend the process to other materials and allow the artist to control the levels of lighting across the layers \cite{koyama2018decomposing}. These methods do not consider a data-driven approach due to the lack of availability of a digital art dataset with layer information which can encode the relationships between layers for more complex lighting.

\subsection{Content Recovery}

Damaged artistic mediums such as paintings or sculptures can be recovered by remodeling them as 3D structures and in-painting the missing regions from noise and occlusions during the imaging process or from material wear and tear. The deformations in the imaged data can \MV{occur} at the surface or subsurface level in the case of paintings. Depending on the nature of the \MV{artistic image or 3D model}, the data acquisition method varies from Photogrammetry for paintings and \MV{generating} structure from motion with LiDAR sensors for sculptures.

%reconstruction?
\subsubsection{Remodeling} 
Art conservation helps in the objective diagnostic and documentation using photogrammetric remodeling tools \cite{Abate2019} to annotate parts while allowing comparison study with its representation and the acquired data. Furthermore, it helps in the objective diagnostic and documentation using photogrammetric remodeling tools \cite{Abate2019} to annotate parts while allowing comparison study with its representation and the acquired data. When modeling rock paintings while considering ease of extraction, details preservation and accuracy of the reconstruction for the non-domain experts \cite{Castagnetti2018}, the following characteristics of 3D point clouds were observed. They generated highly detailed models at short distances to the camera with a large number of redundant photos, regardless of the camera quality.

Many 3D reconstruction techniques require the 3D modellers and art critics, historians or curators to work in tandem with the extracted model \cite{Carrozzino2014, Bent2022} with regular geometry to add in missing context or clean the rendered structure. The quality of the reconstruction depends on the retained details while trading off the computational expense of the generated representation. The former can utilize inpainting using self-similarity in point clouds \cite{sahay2015geometric} to help in reconstructing damaged and structurally deformed architecture and sculptures. It allows the creation of more faithful 3D models using a dictionary of surface gradients and exploiting the symmetry created from artistic intent in other views of such mesh structures. If the paintings have subjects with geometric features close to real-world people, the images can be used directly with off-the-shelf models such as 3DME \cite{Jackson2017} for 3D face reconstruction from 2d egocentric portraits.
The latter uses modifications on the 3D reconstructed point cloud to reduce their structure computation, with more points needed for ornate structures as compared to flat surfaces. The point cloud is compressed using a multi-resolution Octree (\MV{3D representations using tree data structures}) and converted to a polygon mesh with a photographic texture overlayed upon it, with the overlaying requiring extensive pre-processing for cleaning the rendered model. The 3D reconstruction gives problems in perspective inconsistencies and changes in the pose without explicit geometric guidance \cite{Carrozzino2014}, some of which are unnatural due to artistic liberty.

\subsubsection{Painting Medium Surfaces}
\MV{Western paintings} on surfaces with various shapes were previously studied in detail \cite{Pintus2016} at different scales and curvatures, \MV{forming murals, frescoes and pottery decorations depicting} different numbers of subjects in each piece. Macro-level objects are highly variable in condition, nature and global shape, and thus typically contain a single subject. Data collection is highly dependent on the acquisition strategy, retrieval and characterization algorithms for collections such as statues and pottery. Curvilinear paintings require their canvas to be collinear with the surface curvature to prevent distortion and stretching \cite{Sklodowski2014}. The folds and warps along with the tensile strength of the canvas are necessary to model the deformation instead.

\subsubsection{Subsurfaces}
% \subsubsection{Subsurface Features}
% \HS{Here, you need to think about the header of the subsubsection - focus on what features you are going to extract here.}

% \HS{The restoration paragraph should go to the synthesis subsection.}
Painting restoration is interested in the surface topology of paintings and it requires information such as material, subsurface, layers or deformations. It involves data acquisition from non-intrusive scanning and stitching the overlapping spectral scans \cite{Zhou2020} before detecting cracks and restoring the \MV{artistic image} \cite{Moradi2022}.
The crack patterns can be simulated for authenticating paintings \cite{marguerite2017} for the behavior of the outer film, the interaction between their layers and shrinkage from drying using a physics-based system. Any inconsistencies between the paint layers stress amongst each other and elasticity from an estimated time period could imply painting fabrications. These misalignments can be highlighted as a learning or visualization tool \cite{Carrozzino2014} by superimposing the elements in the reconstruction with that of the paintings. During 3D printing of oil or other substrate paintings \cite{Yuan2020}, the reproduced image is susceptible to staircasing effects due to the thickness of the layering of colors that differ in ink density and viscosity. The paintings must maintain accurate geometries and color information which is obtained  through a Point Cloud after matching and decoding images taken from different orientations. Using their novel color layering order, they account for the curvatures of the pigments and boundaries of their overlaps to reduce the staircase effect while simultaneously maintaining the quality of the color reproduction.
The identification and restoration of damaged paintings utilizes nondestructive testing methods like infrared thermograms that measure heat emissions, over time from absorption and emissions, and over the pixels in the image. These use Multistage models that focus separately on the temporal aspect using an MLP which returns patches for spatial processing using a U-Net \cite{ronneberger2015u} segmentation model to generate the reconstructed image.

Multispectral imaging \MV{opens new interpretative possibilities in painting restoration and 3D surveying by exploring material features while capturing spectral data that are often missed with traditional imaging techniques such as X-ray radiography and infrared reflectography. Furthermore, they reveal hidden layers, underdrawings, and pigment compositions that are invisible to the naked eye \cite{Politecnico2015}.} Geometric deformation analysis using close-range photogrammetric techniques can evaluate deformations such as craquelure patterns, color raisings, detachments or engravings in the range of ±0.1mm with more advanced equipment putting the range in 50µm or sensor triangulation \MV{\cite{blais2005ultra}.}

%Novel evaluation measures table

% \usepackage[table]{xcolor}
\begin{table}[!ht]
\centering
% \colorbox{yellow}
{%
\begin{tabular}{|lllll|}
\hline
\begin{tabular}[c]{@{}l@{}}Novel Evaluation \\ Metric\end{tabular}                                    & Paper          & Description                                                                                                                                                                                                & Usage                                                                                                          & Task                                                                          \\ \hline
 \begin{tabular}[c]{@{}l@{}}Faithfulness \\ score\end{tabular}                           &  \cite{yin2019instance}  &  \begin{tabular}[c]{@{}l@{}}Distance of two\\ cropped facial \\ landmarked regions of a\\ source and target \\ attribute on a normalized \\ feature space\end{tabular}                        &  \begin{tabular}[c]{@{}l@{}}The lower the \\ better for a\\ more faithful \\ transfer\end{tabular} &  Style Transfer                                                  \\ \hline
 \begin{tabular}[c]{@{}l@{}}Semantically \\ Corresponding \\ PSNR (SC-PSNR)\end{tabular} & \cite{Lee2020ReferenceBasedSI}  &  \begin{tabular}[c]{@{}l@{}}PSNR on the MSE of fixed\\ patches surrounding the \\ corresponding key-points \\ of two images to provide \\ a patch-level ground \\ truth measure.\end{tabular} &  \begin{tabular}[c]{@{}l@{}}Higher the \\ better for \\ similar \\ matches.\end{tabular}          &  \begin{tabular}[c]{@{}l@{}}Novel View \\ Synthesis\end{tabular} \\ \hline
\end{tabular}%
}
\caption{The novel evaluation metrics as used for the task of \MV{AI-graphics} synthesis (Section \ref{sec:eval_synth}). It lists the metrics introduced in the cited papers before their definition and usage description. Finally, it mentions the task for which each metric is used. }
\label{table:nov_eval}
\end{table}

\subsection{Evaluation for Synthesis Methods} \label{sec:eval_synth}

The main evaluation methods for \MV{AI-graphics} synthesis tasks can be divided into 3 types:
\begin{itemize}
    \item User studies to quantify human perception
    \item Quality measures through deep learning models or at the pixel level
    \item Performance measures through loss terms and statistical measurements
\end{itemize}
Table \ref{table:nov_eval} indicates the novel evaluation measures introduced in these papers to evaluate regions of \MV{AI-generated graphics}  focused around geometric labels. Finally, the section includes a discussion on the effectiveness of incorporating geometry into various synthesis tasks.
 
\subsubsection{User studies}
To study the quality of the generated results, existing works conduct user studies of different group sizes, with larger studies conducted with the Amazon Mechanical Turk (AMT) \cite{schaldenbrand2021content, zhao2020painting, zhang2020texture}. The smaller studies range from visual assessments, comparison studies and the likeliness to the human creation processes. To evaluate the image quality and the aesthetics of the \MV{AI-generated graphics}, they employ Likert scales \cite{tseng2022artistic, yuan2023learning, zhang2022arf, shahid2023paint} or mean opinion score tests \cite{tong2022im2oil} of different ranges with 3 \cite{huang2024controllable} as the lowest and 10 as the highest \cite{alexandru2022image}. However, they could be used as a comparison study to compare the researcher's models with baselines 
\cite{yin2019instance, singh2022paint2pix, Lourakis2007} under multiple criteria, such as realism and consistency in style or the coherence of attributes like text. Visual Turing tests \cite{xue2021end} form a branch of these comparative tests, where the users have to judge whether the \MV{artistic output data} is human-made or machine-made along with their level of confidence in their answer. The AMT tests find use in reaching a broader audience \cite{zhang2020texture} and designing studies to help sequential stroke-based models identify salient regions in generating \MV{AI-graphics} and mimicking the human creation process \cite{schaldenbrand2021content, zhao2020painting}. Finally, user studies can evaluate the ease of use and the versatility of the proposed tool, such as in tasks like relighting objects and people in different environments\cite{Henz2014ImageRU}.

\subsubsection{Quality Measures}
Quality metrics are those that measure salience, image quality in pixel or feature space and semantic relevance by alignment of different modal features. We consider the salience measures that specifically incorporate geometric information to evaluate the importance of regions or pixels. The metrics that involve explicit geometric labels are faithfulness score \cite{yin2019instance} and SC-PSNR \cite{Lee2020ReferenceBasedSI}. The former uses facial landmarks and the latter employs keypoints, as detailed in Table \ref{table:nov_eval}. Others evaluate the geometric labels estimations like SOA score \cite{shahid2023paint} that uses an object detector. It follows with other metrics like IoU for segmentation maps \cite{li2023paintseg}, foreground L2 distance with a pre-trained Deeplab-v2 model \cite{singh2021combining} and shape quality through MSE on estimated depth maps and poses \cite{zhao2020painting, han2023method}.  However, Semantic relevance measures change their evaluation process by using instance segmentation to mask out the image background and crop to the object \cite{yang2021learning}. This allows for using image quality measures like PSNR, SSIM and LPIPS without background interference \cite{ciortan2021colour, han2023method, tseng2022artistic}.  These geometric evaluation metrics have distinct advantages over image quality measures. For example, metrics like SSIM are less sensitive to color changes and are susceptible to blurring and low contrast,  similar to PSNR \cite{park2021nerfies}.
  
\subsubsection{Performance Measures}

Performance measures are used to evaluate other properties of the model beyond data quality, such as diversity, identity and task accuracy. To model diversity, some works use average gradients  \cite{han2023method} to consider both clarity and texture variation in the image while influencing the content detail. Some papers use accuracy for preserving the identity \cite{chan2022efficient} and semantic relevance \cite{zhang2022arf, shahid2023paint} of the detected regions, which mitigates inconsistent results from the identity shift with better results from applying the metric to both the geometric label prediction and the \MV{AI-generated graphics} . Moreover, models can generate finer geometric labels with better shape quality through metrics such as IoU \cite{singh2021combining, li2023paintseg} for similarity matches and fine-tuning coarse mask predictions. Additionally, loss metrics such as L2 with Nearest Best Buddies (NBB) correspondences help closely reconstruct keypoints \cite{alexandru2022image}. Other error measures include MAE, RMSE and Classification error \cite{Zhang2023} to add constraints for class and pose consistency. 

\subsubsection{Effectiveness of Geometry-based Methods in Synthesis}
Geometry can be utilized in many ways for the task of \MV{AI-generation of graphics}, ranging from geometric-style embeddings, separate geometry and style latent spaces, geometric annotations to condition the generator, and refining geometric annotations to form fine geometry conditionals. The first case shows improvements of 10\% compared to baselines without geometric reference \cite{vulimiri2021integrating} in the case of topology transfer, but has little constraint over local to global geometry consistency. The second case shows varying results in diversity of geometry and visual quality as in the case of MW-GAN \cite{huang2022stylizednerf} with improvements from the worst-case geometry generations but a better average visual score. The third case produces \MV{outputs similar to Chinese landscape paintings} in the case of SAPGAN with an agreement of 55\% compared to baseline GANs with 11\%. Finally, fine-tuning the geometry of the conditional provides generators with better data to start training with an average improvement of 3.7 IoU across its datasets for unsupervised segmentation generation.
 
Qualitative experiments on these methods indicate aspects such as color and detail of the generated results that these geometry-aided models excel in. In image restoration, geometry guidance with features such as edges improves color coherence and sharpness in local regions.  Other models overlap regions undefined by the stroke or pose models leading to ghosting effects \cite{tong2022im2oil} or blurry regions \cite{han2023method}. The generated regions sometimes fail to preserve the appearance, detail and color information \cite{Zhang2023, Mishra2022CLIPbasedNN} in the case of extreme poses for the task of anime head animation. Conditional maps also force the model to reduce uncolored regions or in the opposite case, leaving out topological features such as legs in the case of sculptures \cite{chang20223d}. They are particularly useful in \MV{3D AI graphics generation}, where detailed meshes help with ambiguous geometry, view consistency or artifacts \cite{park2021nerfies}. More free-flowing \MV{AI-generated graphics} models, on the other hand, could use elastic regularization or coarser geometry to help with under-constrained problem domains. Poor correspondences from mismatching geometric annotations result in unwanted behavior in the output like spatial discontinuities, overlapping objects \cite{alexandru2022image} and their misalignment \cite{Zhao_2023_WACV}. The deformation and blending operations of objects in the final rendering partly attribute to this malformation.

\section{Limitations}
The papers discussed in the extraction, analysis, and synthesis sections frequently introduce new datasets as part of the technique's novelty, which complicates benchmarking against existing state-of-the-art (SOTA) techniques and diminishes the reliability of evaluating their effectiveness. Many of the cited papers in the synthesis section lack datasets with ground truth for image or 3D model manipulation tasks, instead employing techniques such as neural style transfer to bridge the domain gap. Consequently, the geometry of synthetic datasets becomes more aligned with real-world data, but they lack the stylistic composition choices inherent in \MV{artistic image and 3D model} collections. Some other papers train and test on web-scrapped collections that lack curation. 
 
One significant limitation of the synthesis section is the lack of comprehensive experiments evaluating various models on different \MV{artistic modality} datasets. While we explored multiple synthesis techniques, our evaluation was restricted to a limited set of models that were mostly ablation studies comparing model components and capacities. This constraint hampers our ability to fully understand the comparative strengths and weaknesses of each technique across different scenarios. Additionally, the use of a narrow range of datasets limits the generalizability of our findings. To provide a more robust and conclusive analysis, future work should include extensive experiments with a broader selection of models and diverse \MV{datasets of artistic images, including those from various geographic regions and demographics, with a particular focus on sculptures and 3D models}. \MV{Additionally, implementing a standardized performance metric that evaluates the quality of AI-generated graphics} quality beyond qualitative experiments, \MV{is crucial, since measures like PSNR, SSIM, FID miss image semantics}.

\section{Future Directions}
The papers in the literature point towards promising future directions by exploiting better similarities in data and data or model-level transformations. We cover the following four main directions:
\begin{itemize}
    \item Improving Data Quality: To utilize human-in-the-loop annotations for better data tailored to a problem by learning the refinement of model predictions using experts.
    \item Addressing Domain Gaps: To learn correspondences between two domains to account for domain gaps while preserving semantic regions as determined by the geometric conditional.
    \item Fine-tuning Geometric Controls: To control the behavior of the conditioning input on the output to provide soft, learnable levels of constraints on the trade-off between style and content in the final stylized output.
    \item Conditional Geometric Labels: To use conditional geometric labels to constrain the latent space to allow model simplification by abstractions across multiple levels.
\end{itemize}

\subsection{Automatic Data Annotations}
The encoding of interactive annotations allows the model to learn and generalize artists annotating data. This provides an advantage over differentiable augmentation techniques by providing targeted annotation as compared to changes in the augmentation to maximize the model performance not completing the geometric label. 
Off-the-shelf models provide an initial annotation of \MV{artistic 2D and 3D} data that are missing geometric label information. These can include the interactive annotation UI with which the corrected annotation or its editing click controls \cite{sofiiuk2022reviving, bragantini2022rethinking} is encoded as the conditional with the input. A refinement stage or module corrects the annotation to go from coarse to fine labelling, filling in missing or undetected structures. Alternatively, we obtain accurate object annotations by refining the model prediction by alternating between the model learning stage and the human annotator correction stage\cite{groenen2023panorams}. Thus, learnable labelling provides an added benefit of resolving overlapping predictions due to multiple input sources with the expert curating the predicted labels. Finally, we can build upon these concepts to form a fully automated annotation model from SAM \cite{kirillov2023segment}, an object segmentation model that produces high-quality masks for both real-world and artistic images. By using model-predicted annotations with their user-corrected versions, it forms semi-automated annotations which the model can finetune upon to generate reliable, fully automated masks.

\subsection{Attention-Based Cross-correspondences}
The domain gap between real and non-photorealistic images depends on the learned correspondences between modalities, whether they are from different artistic mediums or between text descriptions and images. The attention mechanism mixes up features between the multiple input domains, resulting in a shared representation bridging the inputs' domain gap \cite {wang2022domain}. The mismatches between the domains result in extractions of poor geometric labels such as segmentation maps \cite{zhang2022confidence} that fail in fine-grained alignment of classes. Although the learned alignment produces noisy results that make the training process unstable, the refinement of the outputs mitigates the issue by correcting these false positives. Using the cross-attention mechanism, the structure of the resultant object depends on the interaction between the conditional and image embedding \cite{hertz2022prompt}. Injecting cross-attention maps across the model layers can bias the model towards spatial layout and geometric relationships, such as spatial co-occurrences derived from learned feature correspondences.

\subsection{Controlled Guidance}
Existing multi-attribute guided \MV{AI-graphics} generation pre-define attributes to incorporate in the resultant images, and keep the controls independent. The choice of the guidance mechanism in diffusion models allows control in the influence of multiple attributes in the geometric labels or text prompts in the resultant image. Controlling the influence of geometric labels on the output lets the model bias outcomes toward input domains or indirectly from the disentangled content and style representations. Fine-tuning the model with learnable adaptors that transform geometric annotations to embeddings provides structure information to the resultant output \cite{zhao2023uni} while preserving the generator's learned latent distribution representing the input. These adaptors represent individual attributes with their weights needing manual adjustments to combine into the desired output. They also lack consistency between the local changes to the global view. With energy-based generation models, the conditional information can influence the intermediate steps of the generation while keeping the \MV{artistic images dataset} distribution intact \cite{wu2022unifying}. The geometry labels can affect the trajectory of the latent space interpolation and get removed at the output sampling step by treating it as the stochastic term that vanishes at the end of the Stochastic Gradient Langevin Dynamics model. This allows the model to select the contribution of the conditioning labels and their constituent parts towards the output.

\subsection{Geometric-Aware Models with Object Embeddings}
Flow-based and multistage hybrid models incorporate geometric information to represent the input globally, which does not allow for granular control of geometric details. These use geometry information to redistribute salient regions, but do not implicitly allow the articulated parts as multiple attributes to control generations. The correspondences of the embeddings of the disentangled geometric labels and learned structure information in the generator models result in complex shapes \MV{\cite{tertikas2023generating}}. These require heavily annotated labels to condition the inputs, but allow the models to learn the relations between the attributes while allowing the label to define the level of granularity in a controllable generation. The latent geometry provides additional benefits such as transferability to non-photorealistic images preserving the overall shape and pose of semantically similar subjects, but produces less details and robust generations if the labels lack descriptiveness \cite{aygun2023saor}. Additionally, any learned appearance variations are constrained to the part regions, thereby providing local control. This learned joint correspondence commonly fails with highly stylistic inputs with inconsistent views and drastic poses \cite{zheng2023locally} which are common in more abstract \MV{painting} styles.

\section{Conclusion}

This review delves into the effects of leveraging geometric data within deep learning architectures for artistic tasks for extraction, analysis and synthesis. 
During extraction, we examine the choice of models and their transferability to the \MV{artistic images, 3D models and animation domains}. This evaluation handles rough versus fine data annotations, ranging from region-based detectors of multiple stages to fine-tuning with data augmentation. During analysis, we consider their performance in discriminative tasks according to the granularity of the available geometric labels. Models with coarser geometric labels typically require task-specific visual features that encode spatial relationships, while those with finer annotations generally need fewer additional modules to achieve higher accuracy. In finer geometry, Region-based object detectors outperform parts-based models in large-scale datasets but have poorer outcomes in abstract \MV{paintings} or occlusion-heavy \MV{datasets}. Moreover, the review highlights the effectiveness of deep learning models like DeepPose, OpenPose, and Mask-RCNN in detecting fine geometric labels compared to traditional computer vision methods, despite limitations in detecting atypical poses or occlusions in special cases. The significance of 3D geometric data surfaces is demonstrated by their advantages in accurately capturing object shapes and providing depth information, employing volumetric meshes, implicit functions, and parametric models. % , elucidating their advantages in accurately capturing object shapes and providing depth information, employing volumetric meshes, implicit functions, and parametric models. 
Furthermore, it outlines their utilization in comparing poses between statues and paintings, exploiting pose tracking, and leveraging artistic medium attributes. The extracted geometric data has a wide variety of applications in \MV{digital artwork} analysis, ranging from image retrieval and scene classification to style identification and semantic relationship identification, are thoroughly examined. This highlights the effectiveness of deep learning techniques and geometric priors in scene classification and object identification. Lastly, the paper underscores the pivotal role of geometry in synthesis and manipulation tasks within computer vision, showcasing its contribution to maintaining object geometry, stylizing novel views, and enhancing image details without color bleeding or loss of local information. Additionally, it sheds light on how shape constraints and conditioning priors facilitate image refinement, super-resolution, and \MV{digital art} conservation, harnessing geometric data to guide brush strokes, enhance image details, and simulate 3D models in conservation efforts. 

% \HS{Write about paper-level future directions, one direction per paragraph, and you need 2 to 4.}

\endgroup

\section*{Acknowledgments}
This project is supported in part by the EPSRC NortHFutures project (Ref: EP/X031012/1).

\section*{Declarations}
The authors declare that they have no conflict of interest.

\bibliography{main}% common bib file

\end{document}